\documentclass[
  a4paper,        
  techreport,     
]{art}
\usepackage{cmap}
\usepackage{graphicx}
\usepackage{wrapfig}
\usepackage{tcolorbox}
\usepackage{xcolor}
\usepackage{enumitem}
\usepackage{fontawesome} 
\usepackage{fvextra}  
\usepackage{geometry}
\usepackage{longtable}
\usepackage{booktabs}
\usepackage{multirow}
\usepackage{colortbl}
\usepackage{siunitx}

\definecolor{acad}{RGB}{235,244,250}     
\definecolor{busi}{RGB}{235,245,235}     
\definecolor{tech}{RGB}{242,240,247}     
\definecolor{heal}{RGB}{252,240,240}     
\definecolor{life}{RGB}{252,250,232}     
\definecolor{home}{RGB}{252,245,235}     
\definecolor{cult}{RGB}{232,247,247}     
\definecolor{envi}{RGB}{245,245,237}     
\definecolor{misc}{RGB}{245,245,245}     

\definecolor{highrelevance}{HTML}{0EA5E9}  
\definecolor{mediumrelevance}{HTML}{38BDF8}  
\definecolor{lowmediumrelevance}{HTML}{7DD3FC}  
\definecolor{lowrelevance}{HTML}{E0F2FE}  

\definecolor{boxbg}{RGB}{248, 249, 250}
\definecolor{bordercol}{RGB}{59, 130, 246}

\definecolor{bg}{rgb}{0.95,0.95,0.95}

\tcbset{
    enhanced,
    colback=boxbg,
    colframe=bordercol,
    arc=5mm,
    boxrule=1pt,
    top=10mm,
    bottom=10mm,
    left=10mm,
    right=10mm,
    boxsep=10mm,
    drop fuzzy shadow
}

\newcommand{\pubtitle}{Mobile-MMLU: A Mobile Intelligence Language Understanding Benchmark}

\newcommand{\pubauthA}{Sondos Mahmoud Bsharat}
\newcommand{\pubaffilA}{a}
\newcommand{\eqcontribA}{}

\newcommand{\pubauthB}{Mukul Ranjan}
\newcommand{\pubaffilB}{a}
\newcommand{\eqcontribB}{}

\newcommand{\pubauthC}{Aidar Myrzakhan}
\newcommand{\pubaffilC}{a}
\newcommand{\eqcontribC}{}

\newcommand{\pubauthD}{Jiacheng Liu}
\newcommand{\pubaffilD}{a}

\newcommand{\pubauthE}{Bowei Guo}
\newcommand{\pubaffilE}{a}

\newcommand{\pubauthF}{Shengkun Tang}
\newcommand{\pubaffilF}{a}

\newcommand{\pubauthG}{Zhuang Liu}
\newcommand{\pubaffilG}{b}

\newcommand{\pubauthH}{Yuanzhi Li}
\newcommand{\pubaffilH}{c}

\newcommand{\pubauthI}{Zhiqiang Shen$^{\dag}$}
\newcommand{\pubaffilI}{a}
\newcommand{\authemailA}{zhiqiang.shen@mbzuai.ac.ae}

\newcommand{\pubaddrA}{VILA Lab, MBZUAI}
\newcommand{\pubaddrB}{Princeton University}
\newcommand{\pubaddrC}{Apple}

\newcommand{\pubemail}{\authemailA}

\newcommand{\pubabstract}{
Rapid advancements in large language models (LLMs) have increased interest in deploying them on mobile devices for on-device AI applications. Mobile users interact differently with LLMs compared to desktop users, creating unique expectations and data biases. Current benchmark datasets primarily target at server and desktop environments, and there is a notable lack of extensive datasets specifically designed for mobile contexts. Additionally, mobile devices face strict limitations in storage and computing resources, constraining model size and capabilities, thus requiring optimized efficiency and prioritized knowledge. To address these challenges, we introduce \texttt{\bf Mobile-MMLU}, a large-scale benchmark dataset tailored for mobile intelligence. It consists of 16,186 questions across 80 mobile-related fields, designed to evaluate LLM performance in realistic mobile scenarios. A challenging subset, \texttt{\bf Mobile-MMLU-Pro}, provides advanced evaluation similar in size to MMLU-Pro but significantly more difficult than our standard full set. Both benchmarks use {\em multiple-choice}, {\em order-invariant} questions focused on practical mobile interactions, such as recipe suggestions, travel planning, and essential daily tasks. The dataset emphasizes critical mobile-specific metrics like inference latency, energy consumption, memory usage, and response quality, offering comprehensive insights into model performance under mobile constraints. Moreover, it prioritizes privacy and adaptability, assessing models' ability to perform on-device processing, maintain user privacy, and adapt to personalized usage patterns. \texttt{\bf Mobile-MMLU} family offers a standardized framework for developing and comparing mobile-optimized LLMs, enabling advancements in productivity and decision-making within mobile computing environments.
Our code and data are available at: \url{https://github.com/VILA-Lab/Mobile-MMLU}.
}

\keywords{
Mobile Intelligence\sep Large Language Models\sep Order-invariant Benchmarking\sep On-Device AI\sep
}

\title{\pubtitle}

\newcommand{\dg}{\textsuperscript{\textbf{\normalfont$*$\textnormal{,}}}}

\ifdef{\pubauthA}{\author[\pubaffilA]{\pubauthA\ifdef{\orcidA}{~\protect\orcid{\orcidA}}{}\ifdef{\eqcontribA}{\dg}{}}}{}
\ifdef{\pubauthB}{\author[\pubaffilB]{\pubauthB\ifdef{\orcidB}{~\protect\orcid{\orcidB}}{}\ifdef{\eqcontribB}{\dg}{}}}{}
\ifdef{\pubauthC}{\author[\pubaffilC]{\pubauthC\ifdef{\orcidC}{~\protect\orcid{\orcidC}}{}\ifdef{\eqcontribC}{\dg}{}}}{}
\ifdef{\pubauthD}{\author[\pubaffilD]{\pubauthD\ifdef{\orcidD}{~\protect\orcid{\orcidD}}{}\ifdef{\eqcontribD}{\dg}{}}}{}
\ifdef{\pubauthE}{\author[\pubaffilE]{\pubauthE\ifdef{\orcidE}{~\protect\orcid{\orcidE}}{}\ifdef{\eqcontribE}{\dg}{}}}{}
\ifdef{\pubauthF}{\author[\pubaffilF]{\pubauthF\ifdef{\orcidF}{~\protect\orcid{\orcidF}}{}\ifdef{\eqcontribF}{\dg}{}}}{}
\ifdef{\pubauthG}{\author[\pubaffilG]{\pubauthG\ifdef{\orcidG}{~\protect\orcid{\orcidG}}{}\ifdef{\eqcontribG}{\dg}{}}}{}
\ifdef{\pubauthH}{\author[\pubaffilH]{\pubauthH\ifdef{\orcidH}{~\protect\orcid{\orcidH}}{}\ifdef{\eqcontribH}{\dg}{}}}{}
\ifdef{\pubauthI}{\author[\pubaffilI]{\pubauthI\ifdef{\orcidI}{~\protect\orcid{\orcidI}}{}\ifdef{\eqcontribI}{\dg}{}}}{}
\ifdef{\pubauthJ}{\author[\pubaffilJ]{\pubauthJ\ifdef{\orcidJ}{~\protect\orcid{\orcidJ}}{}\ifdef{\eqcontribJ}{\dg}{}}}{}
\ifdef{\pubauthK}{\author[\pubaffilK]{\pubauthK\ifdef{\orcidK}{~\protect\orcid{\orcidK}}{}\ifdef{\eqcontribK}{\dg}{}}}{}

\ifdef{\eqcontribA}{\equalcontrib{}}{}
\ifdef{\eqcontribB}{\equalcontrib{}}{}
\ifdef{\eqcontribC}{\equalcontrib{}}{}
\ifdef{\eqcontribD}{\equalcontrib{}}{}
\ifdef{\eqcontribE}{\equalcontrib{}}{}
\ifdef{\eqcontribF}{\equalcontrib{}}{}
\ifdef{\eqcontribG}{\equalcontrib{}}{}
\ifdef{\eqcontribH}{\equalcontrib{}}{}
\ifdef{\eqcontribI}{\equalcontrib{}}{}
\ifdef{\eqcontribJ}{\equalcontrib{}}{}
\ifdef{\eqcontribK}{\equalcontrib{}}{}

\ifdef{\pubaddrA}{\affil[a]{\pubaddrA}}{}
\ifdef{\pubaddrB}{\affil[b]{\pubaddrB}}{}
\ifdef{\pubaddrC}{\affil[c]{\pubaddrC}}{}
\ifdef{\pubaddrD}{\affil[d]{\pubaddrD}}{}
\ifdef{\pubaddrE}{\affil[e]{\pubaddrE}}{}
\ifdef{\pubaddrF}{\affil[f]{\pubaddrF}}{}
\ifdef{\pubaddrG}{\affil[g]{\pubaddrG}}{}
\ifdef{\pubaddrH}{\affil[h]{\pubaddrH}}{}

\contact{\pubemail}

\begin{abstract}
    \pubabstract{}
\end{abstract}

\ifdef{\pubkeywords}{\keywords{\pubkeywords}}{}

\ifdef{\pubVadj}{\verticaladjustment{\pubVadj}}{}

\pubdoi{Technical Report}
\pubdate{\today}

\begin{document}
\maketitle


\section{Introduction}

Over the past decade, the rapid adoption of mobile devices has transformed how people consume information and interact with technology. As mobile hardware becomes increasingly powerful and ubiquitous, there is growing demand for on-device artificial intelligence solutions capable of supporting real-time language understanding. From virtual assistants to language translation apps, mobile language intelligence is reshaping communication, learning, and productivity at unprecedented scales. Despite these advancements, there remains a pressing need for benchmarks that can rigorously evaluate language understanding models specifically designed for mobile environments. Tech companies like Apple recently introduces initiatives of {\em Apple Intelligence}~\cite{gunter2024apple}. However, their used evaluation datasets or benchmarks remain focused on desktop and server-side use cases, creating a gap in the ability to assess and optimize LLMs for on-device applications. 

\begin{figure}[t]
    \centering
    \begin{subfigure}[b]{0.48\textwidth}
        \centering
        \includegraphics[width=\textwidth]{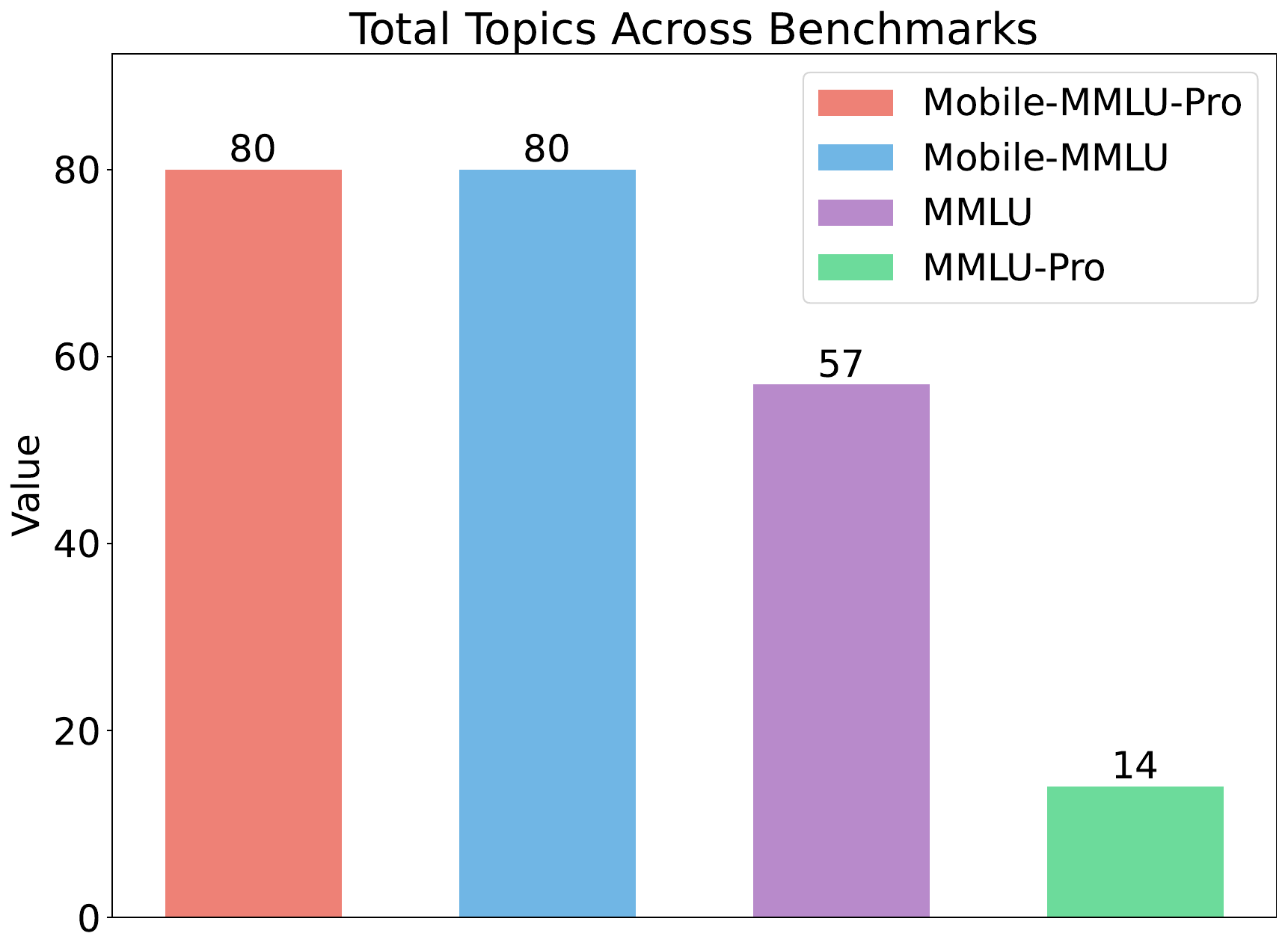}
        \caption{Number of topics across benchmark}
        \label{fig:topics_bench}
    \end{subfigure}
    \hfill
    \begin{subfigure}[b]{0.48\textwidth}
        \centering
        \includegraphics[width=\textwidth]{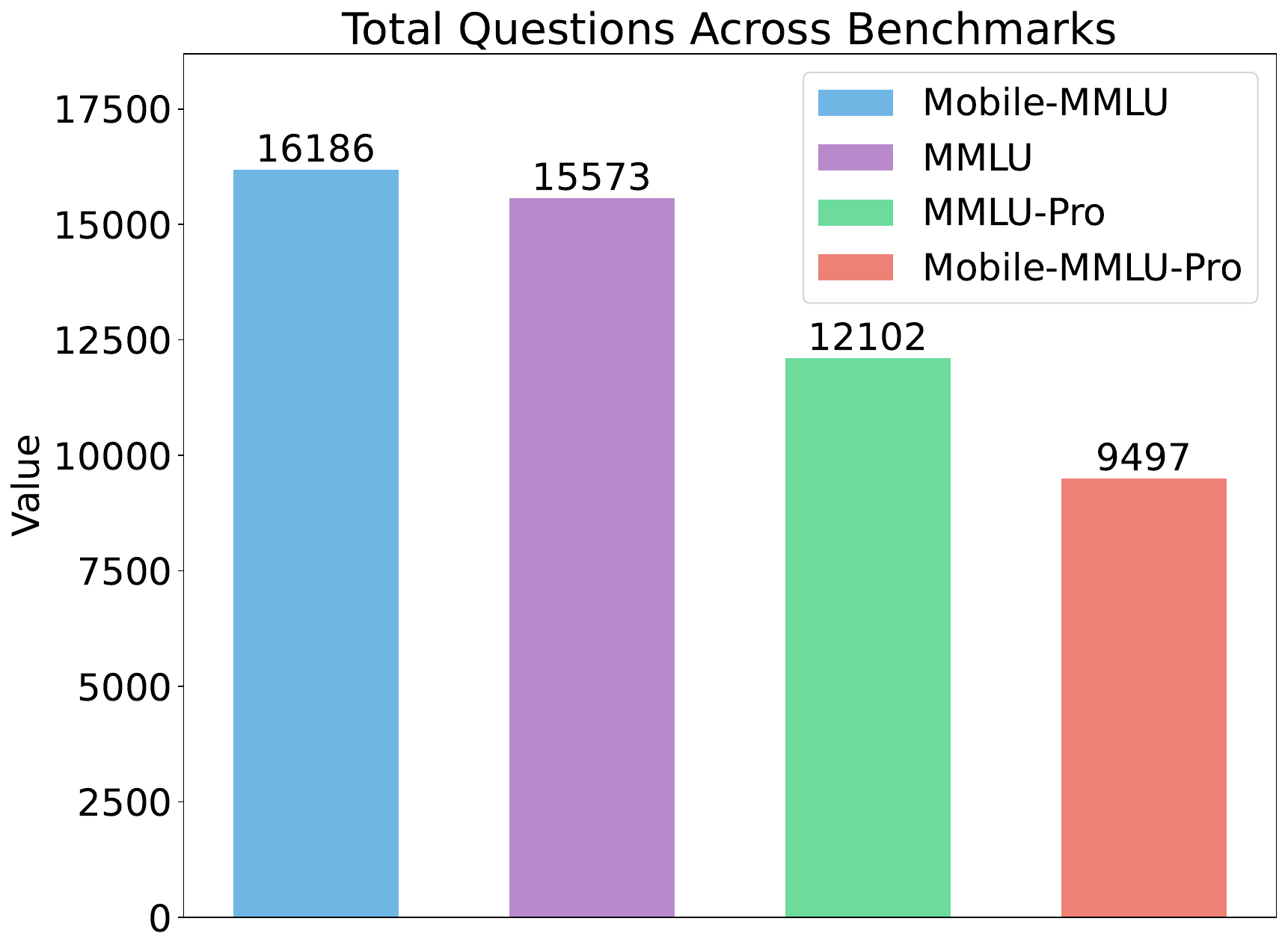}
        \caption{Number of questions across benchmark}
        \label{fig:questions_bench}
    \end{subfigure}
    \caption{Statistical comparison across benchmarks. Our Mobile-MMLU and Mobile-MMLU-Pro both maintain comprehensive coverage with 80 topics each, significantly more than MMLU (57 topics) and MMLU-Pro (14 topics). In terms of question volume, Mobile-MMLU leads with 16,186 questions, followed by MMLU (15,573), MMLU-Pro (12,102), and Mobile-MMLU-Pro (9,497). Our full version contains the largest number of questions, designed for the mobile-centric evaluation of mobile-level LLMs. Oue Pro version has fewer but more challenging questions, making it ideal for quick testing of strong models.} 
    \label{fig:topic_dist}
    \vspace{0.1in}
\end{figure}

Generally, these existing large-scale desktop or cloud-based systems-level language understanding benchmarks usually fail to capture the unique constraints and performance targets required for mobile devices. Concerns like limited memory, strict power budgets, and real-time inference demands mean that a model’s performance in the cloud may not translate well to the phone in your pocket. Furthermore, mobile devices exhibit diverse operating conditions, varying hardware capabilities, network connectivity, and sensor inputs, which complicate attempts to directly apply traditional benchmarks.

To address these challenges, we introduce Mobile-MMLU, a comprehensive benchmark designed to evaluate LLM performance in mobile-specific contexts. Mobile-MMLU spans 80 diverse mobile-related domains, containing over 16,000 questions carefully curated to reflect real-world mobile usage patterns. {\em By focusing on tasks that align with typical mobile interactions, Mobile-MMLU ensures that models are tested in scenarios that matter most to mobile users.} Moreover, our benchmark provides a standardized framework to measure mobile-specific and order-invariant\footnote{Our {\em order-invariance} is demonstrated through two key aspects: 1) The lengths of correct and incorrect options are maintained consistently, with the incorrect options sometimes being longer than the correct ones to prevent bias and favor in selection by the small-scale LLMs; and 2) The results remain consistent with accuracy obtained from a random-order arrangement scheme.} performance metrics, enabling developers to optimize LLMs for the unique constraints of mobile computing.

Furthermore, currently testing results often vary across multiple evaluations of the same LLM or among different strong LLMs. To address these inconsistencies and increase benchmark difficulty, we propose creating a model-consistent version of Mobile-MMLU-Pro derived from a carefully filtered subset of Mobile-MMLU (the full set). This scheme reduces test samples, lowers computational demands, and excludes questions that yield inconsistent predictions across various LLMs. Consequently, this strategy enhances test difficulty and ensures better reliability and consistency of the evaluation process.
A key aspect of Mobile-MMLU and Mobile-MMLU-Pro is its focus on privacy and personalization. Mobile devices often handle sensitive user data, making on-device processing a critical requirement for preserving user privacy. Our benchmark evaluates models' ability to operate efficiently on-device while adapting to various user behaviors and preferences. This specialized property ensures that the evaluated models are not only technically efficient but also capable of providing meaningful and secure interactions for mobile users.

Mobile-MMLU/-Pro represents a significant step forward in the development and evaluation of mobile-optimized LLMs. By providing a robust and comprehensive framework, it empowers researchers and developers to address the unique challenges of mobile platforms. In sum, Mobile-MMLU and Mobile-MMLU-Pro set out to bridge the gap between state-of-the-art language models and the demands of mobile ecosystems. By providing a standardized, realistic, and rigorous testing suite, our benchmark enables the community to build, evaluate, and refine mobile-centric natural language understanding models. We envision that Mobile-MMLU and Mobile-MMLU-Pro will serve as a valuable stepping stone for researchers looking to optimize the performance, efficiency, and reliability of language understanding technologies on the ever-growing landscape of mobile platforms.

\begin{figure}[t]
    \centering
    \includegraphics[width=0.99\textwidth]{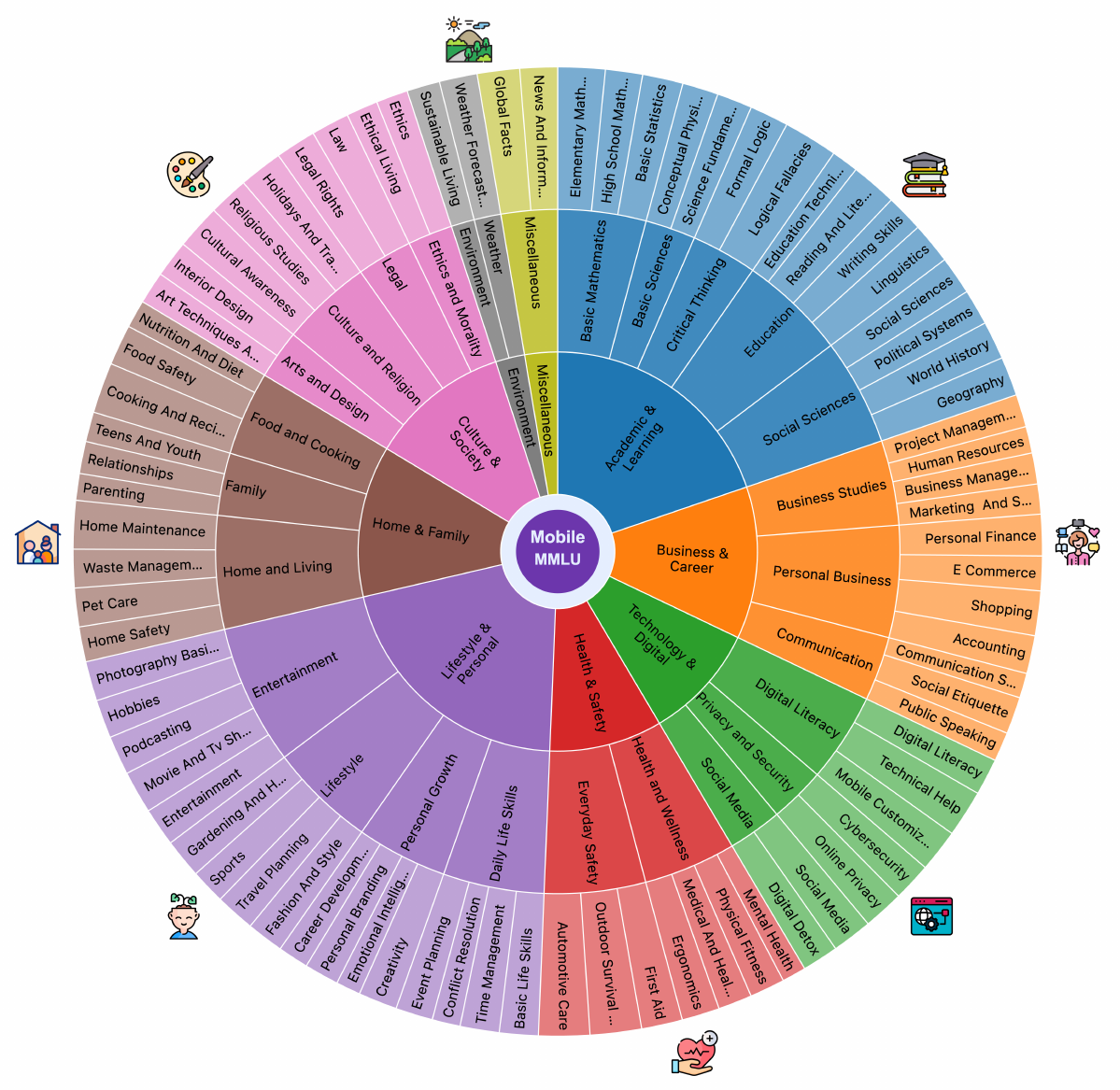}
    \caption{Illustration of topic hierarchy of Mobile-MMLU. Our benchmark consists of topics and questions related to daily life use cases like \texttt{Travel-Planning}, \texttt{First-Aid}, \texttt{Parenting}, etc.} 
    \label{fig:hierarchy}
    \vspace{0.15in}
\end{figure}

\section{Related Work}
The evaluation and deployment of Large Language Models (LLMs) on mobile and edge devices has emerged as a critical research direction, driven by demands for enhanced privacy, reduced latency, and personalized user experience \cite{pan2020privacy, wang2024llms, buesser2024private, xu2024device, zhang2024mobile}. However, few benchmarks have been specifically developed for mobile scenarios and mobile LLMs.

MMLU \cite{hendrycks2020measuring} serves as the foundational evaluation benchmark for most of the LLM, assessing their capabilities across different domains including STEM, humanities, social sciences. MMLU-Pro  \cite{wang2024mmlu}, extending this framework by incorporated reasoning-focused questions and expanding the choice set from four to ten options. Benchmarks including HELM \cite{liang2022holistic}, GLUE \cite{wang2018glue}, SuperGLUE \cite{wang2019superglue}, BigBench \cite{srivastava2022beyond}, HellaSwag \cite{zellers2019hellaswag}, GPQA \cite{rein2023gpqa}, and ARC \cite{clark2018think} have further contributed to evaluating LLMs' generalization and reasoning capabilities. Platforms such as OpenCompass \cite{2023opencompass}, Chatbot Arena \cite{chiang2024chatbot}, and Open-LLM-Leaderboard \cite{myrzakhan2024open} have standardized evaluation methodologies and enabled transparent model comparisons. However, these benchmarks are primarily designed for evaluating general understanding of language models in desktop environments, overlooking the unique constraints and use cases specific to mobile platforms. On the other hand, most of the mobile-device specific benchmarks primarily focus on technical performance metrics, overlooking the distinct nature of mobile interactions \cite{wang2024hammerbench, murthy2024mobileaibench, li2024palmbench}.  HammerBench \cite{wang2024hammerbench} focuses on evaluating function-calling capabilities in real-world mobile scenarios, particularly analyzing how LLMs handle imperfect instructions and context shifts in human-LLM interactions. MobileAIBench \cite{murthy2024mobileaibench} provides a comprehensive framework for assessing both LLMs and LMMs across different model sizes and quantization levels, with particular attention to trust, safety, and hardware resource utilization on mobile devices. PalmBench \cite{li2024palmbench} specifically targets the evaluation of compressed models on mobile platforms, examining the critical balance between generative performance, latency, and resource efficiency.

While these benchmarks have made significant progress in evaluating specific aspects of mobile LLM deployment, there remains a need for a unified benchmark that comprehensively assesses both the technical performance and real-world utility of LLMs in mobile-specific scenarios. Recent studies on mobile information needs \cite{khaokaew2023understanding, guy2016searching, trippas2024users} have shown that mobile users exhibit distinct interaction patterns and requirements compared to desktop users. Our work, Mobile-MMLU family, addresses this gap by providing a holistic evaluation framework that considers both the unique constraints of mobile platforms and the distinct patterns of mobile user interactions by incorporating different aspects of daily life scenarios in our benchmark.

\section{Mobile-MMLU Benchmark}

\subsection{Overview}

Our Mobile-MMLU benchmark series is designed to evaluate language models in mobile-specific contexts with two specialized versions: full Mobile-MMLU (16,186 questions) and Mobile-MMLU-Pro (9,497 questions). Covering 80 practical domains such as {\em First Aid} and {\em Travel Planning}, as illustrated in Figure~\ref{fig:hierarchy}, these datasets prioritize real-world mobile applications for mobile-level LLMs over traditional academic subjects.

Our dataset construction pipeline follows four key stages:
\begin{itemize}
    \item 	{\em Field selection}. We aim to identify mobile-relevant domains using sources like WikiHow, Stack Exchange, Reddit, other forums, and LLM suggestions.
    \item   {\em Question / Choice generation}. We create scenario-based MCQs using GPT-4O \cite{hurst2024gpt}, O1-preview \cite{jaech2024openai}, and human verification.
    \item   {\em Similarity filtering}. We apply MPNet embeddings~\cite{song2020mpnet} to remove duplicates (cosine similarity < 0.98).
    \item   {\em Human-AI collaborative verification}. We further refine through iterative annotation and multi-model consensus validation to remove low-quality or easy question samples.
\end{itemize}

Some observed issues and precautions in our data construction:
\begin{itemize}
\item {\em LLM selection order bias}. To verify whether LLMs have a tendency to prefer certain answer positions, we test model performance by placing the ground truth (GT) answer in different ranking positions.
\item {\em LLM bias toward option length}. We observe that LLMs (especially mobile-level LLMs) tend to select the longest option more frequently. To mitigate this, we adjust the length of incorrect options to be similar to or longer than the GT answer. This ensures that the model is selecting based on knowledge rather than guessing based on option length.
\end{itemize}

\subsection{Benchmark Construction Pipeline}

\begin{figure}[t]
  \centering
    \includegraphics[width=0.99\linewidth]{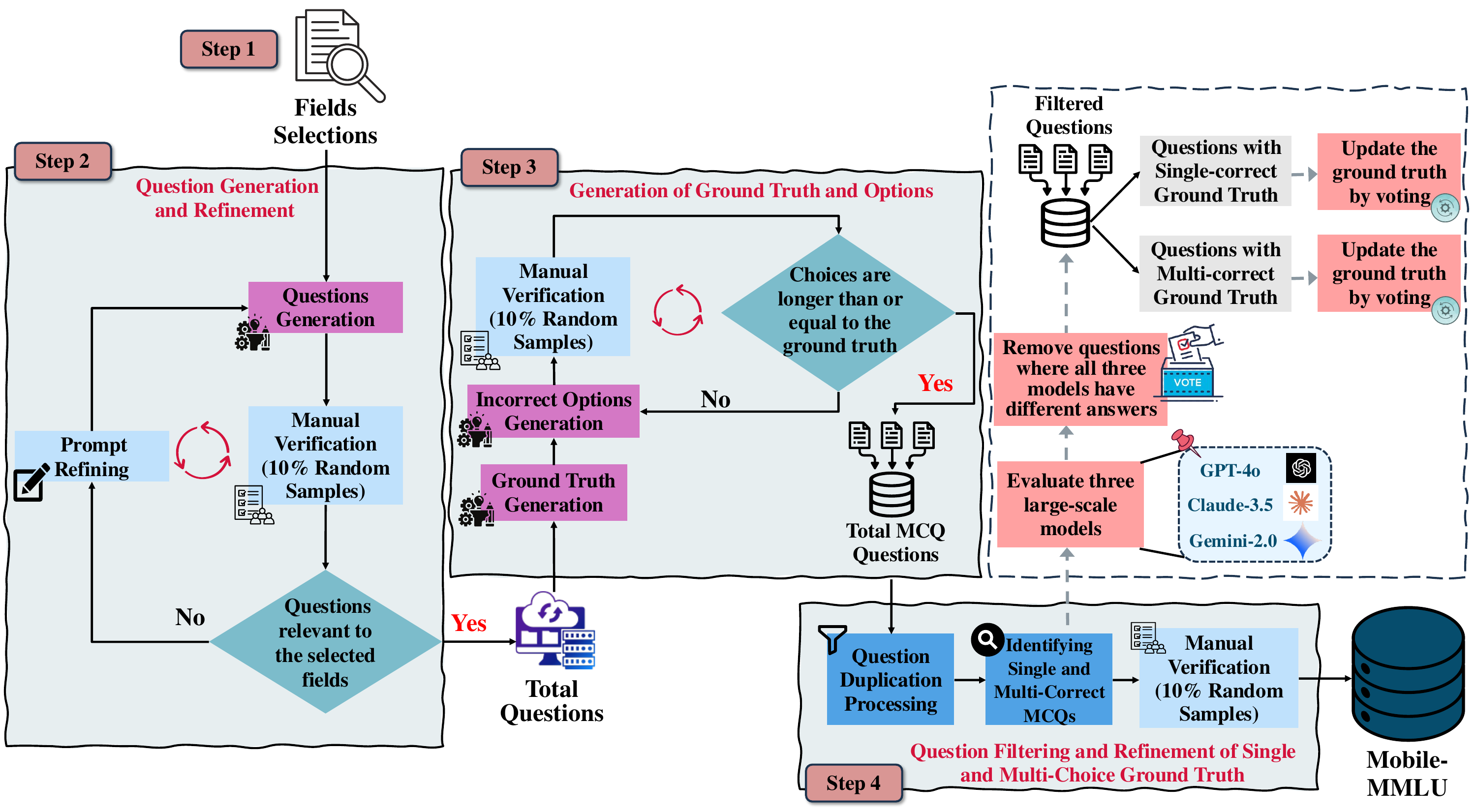}
  \caption{Data construction pipeline for Mobile-MMLU, where the left subfigure (Step 2) illustrates handling ground truth and multiple-choice questions (MCQs), and the right subfigure (Step 4) demonstrates handling multi-correct situations. The whole process includes field selection, question structuring, and iterative verification for mobile-relevant questions.}
  \label{pipeline-final}
  \vspace{0.1in}
\end{figure}

\noindent{\bf Field Selection:} As shown in Figure \ref{pipeline-final}, our benchmark construction begins with a comprehensive search from scratch to identify domains relevant to everyday activities, work, shopping, gaming, travel, and other practical scenarios. The objective is to ensure that the chosen fields are relevant to real-world user needs on mobile devices and align with the types of questions and searches people typically make on their phones. To achieve this, we gather fields from multiple sources, including Wikipedia and various websites. Additionally, to ensure comprehensive coverage, we utilize LLMs to generate suggestions for additional fields. By combining these diverse fields, we ensure that the selected fields are comprehensive, inclusive, and reflective of practical user scenarios.

\noindent{\bf Question Generation:} Users in real life encounter a wide range of situations, from simple, direct queries to complex, multi-step problem-solving scenarios. To reflect this diversity, we design our dataset with two difficulty levels of questions. For each selected field, the initial questions are generated using GPT-4o and o1-preview. The first level is the ordinary questions that involve straightforward scenarios, such as ``How can I check who viewed my LinkedIn profile?'' or ``How can I remove oil stains from my driveway?''; and the second level is the complex, scenario-based questions that require multistep reasoning, decision-making, and the ability to evaluate multiple factors before arriving at a conclusion. This comprises 6,020 questions out of the total. For instance, ``I scheduled a series of posts on Facebook and Instagram using a third-party app, but some posts didn't publish. Considering potential issues like API limitations, platform policies, and app permissions, what might be causing this, and how can I fix it?'' or ``My car's check engine light just turned on, but I need to make a long drive tomorrow. How can I determine if it's safe to drive or if I need immediate repairs?''
These two levels of questions are generated in separate batches, each following the methodology described in Step 2 of Figure \ref{pipeline-final}. We apply two types of prompt formulations for the practical implementation.

\noindent{\bf Generation of Ground Truth and Options:} The next step is to generate the correct (ground truth) answer for each question, followed by generating incorrect options. To avoid LLM bias toward option length, these incorrect options are designed to be either the same length as the correct answer or slightly longer, with differences only in specific keywords. This principle was adopted after human verification revealed that evaluated models tend to prefer longer answers. To mitigate this bias, incorrect options are designed to be equal to or longer than the correct answers while maintaining similar wording and structure. This approach not only addresses length bias but also increases the difficulty of our benchmark.

\noindent{\bf Similarity filtering:} Once the questions with their options are constructed, a cosine similarity metric is applied across all questions to detect and remove duplicates or questions with significant overlap. This step ensures that the dataset consists of unique, non-repetitive questions. To compute cosine similarity, we use vector representations generated by the ``all-mpnet-base-v2'' model \cite{song2020mpnet} from Sentence-Transformer \cite{reimers-2019-sentence-bert}. This model encodes each question into a dense vector of size  768, where each dimension captures semantic and contextual information about the question. The cosine similarity between two questions is then calculated as:
\begin{equation}
\text{Cosine Similarity} = \frac{Q_1 \cdot Q_2}{\|Q_1\| \cdot \|Q_2\|}=\frac{\sum_{k=1}^{n} Q_{1,k} Q_{2,k}}{\sqrt{\sum_{k=1}^{n} Q_{1,k}^2} \cdot \sqrt{\sum_{k=1}^{n} Q_{2,k}^2}}
\end{equation}
where \( Q_1 \) and \( Q_2 \) represent the vectorized forms of the two questions, and \( n = 768 \) corresponds to the embedding dimension.

\subsection{Human Annotation and Refinement}

\noindent{\bf Refinement During the  Generation Process:}
The human annotation process starts during question generation to ensure quality and relevance. It consists of two key phases:

\begin{itemize}
    \item Phase 1: {\em Verification of question relevance}. Human reviewers assess whether the generated questions are relevant to the corresponding field and applicable to mobile use cases. If a selected sample contains questions that are irrelevant or misaligned with mobile use case scenarios, the batch is regenerated through prompt refinement.
    
    \item Phase 2: {\em Validation of incorrect options}.
    Reviewers verify that the incorrect options are indeed incorrect and ensure that they are either longer than or equal in length to the ground truth answers.
\end{itemize}

The above process is repeated for each generated batch.

\noindent {\bf Refinement of Single and Multi-Correct Ground Truth}:
We observe that some questions have multiple correct answers due to the nature of the options generated in Phase 2. To address this, as illustrated in Figure \ref{pipeline-final} (Step 4.2), we analyze responses from three large-scale models, GPT-4o, Claude-3.5, and Gemini-2.0. If all three models provide different answers (i.e., all three models' predictions are inconsistent and no two models select the same option), we remove these questions, as this suggested that multiple options are likely correct, or no absolutely correct choice here. 

\begin{figure}[t]
  \centering
    \includegraphics[width=0.95\linewidth]{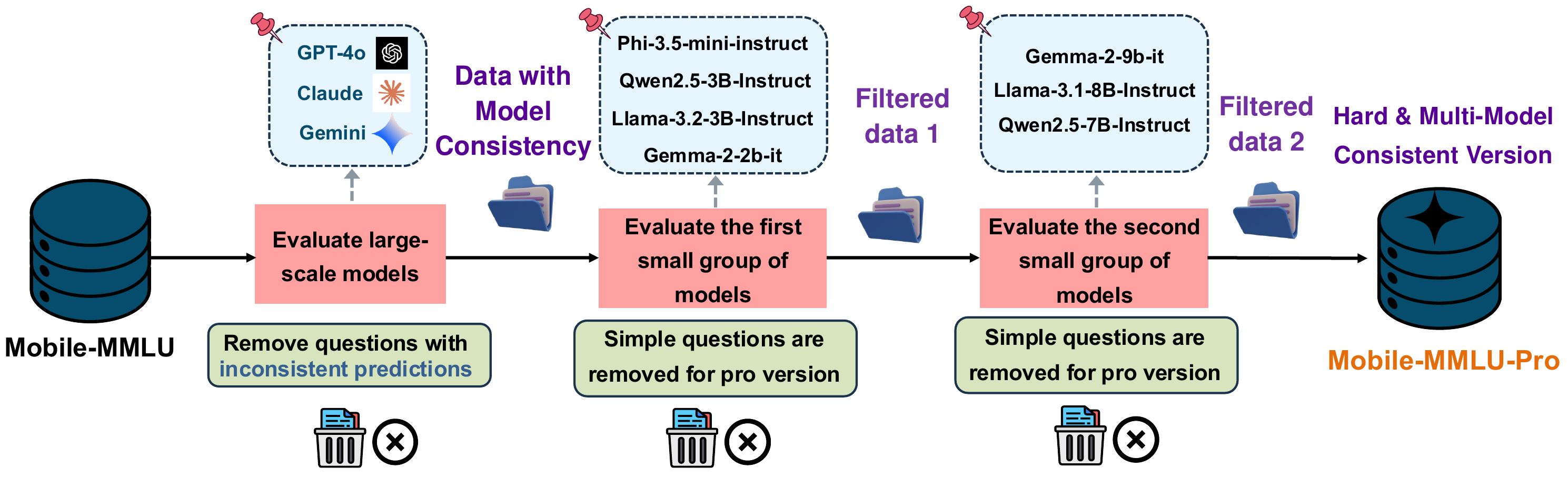}
  \caption{Data creation pipeline for Mobile-MMLU-Pro. }
  \label{pipeline-pro}
  \vspace{0.1in}
\end{figure}

For the remaining questions, we update the ground truth using a voting process, where the updated ground truth corresponds to the answer agreed upon by at least two out of three models. As a result, {\em 5.8\% of the updated questions now have more than one correct choice as the ground truth}\footnote{This means our benchmark contains 5.8\% questions with multi-choice as ground-truth.}. For the rest, if the voting process confirms the initial answer, the ground truth remains unchanged. Otherwise, the most agreed-upon answer is selected as the new ground truth.

\section{Mobile-MMLU-Pro: A Multi-model Consistency and More Challenging Version}

\begin{figure}[t]
  \centering
    \includegraphics[width=1\linewidth]{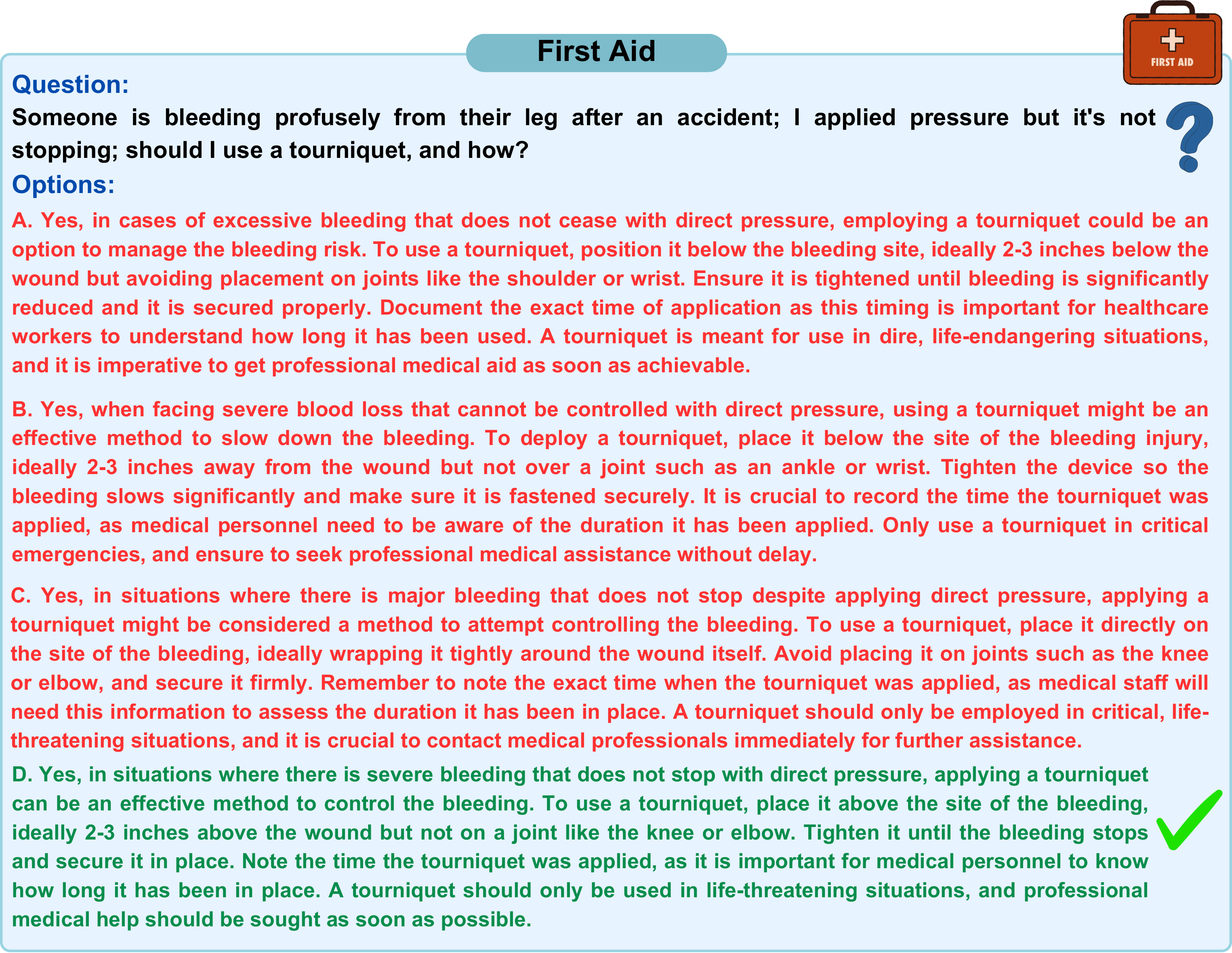}
  \caption{Example question from the {\em First Aid} field in Mobile-MMLU dataset.}
  \label{first_aid_example1}
  \vspace{0.1in}
\end{figure}

\begin{figure}[t]
  \centering
    \includegraphics[width=1\linewidth]{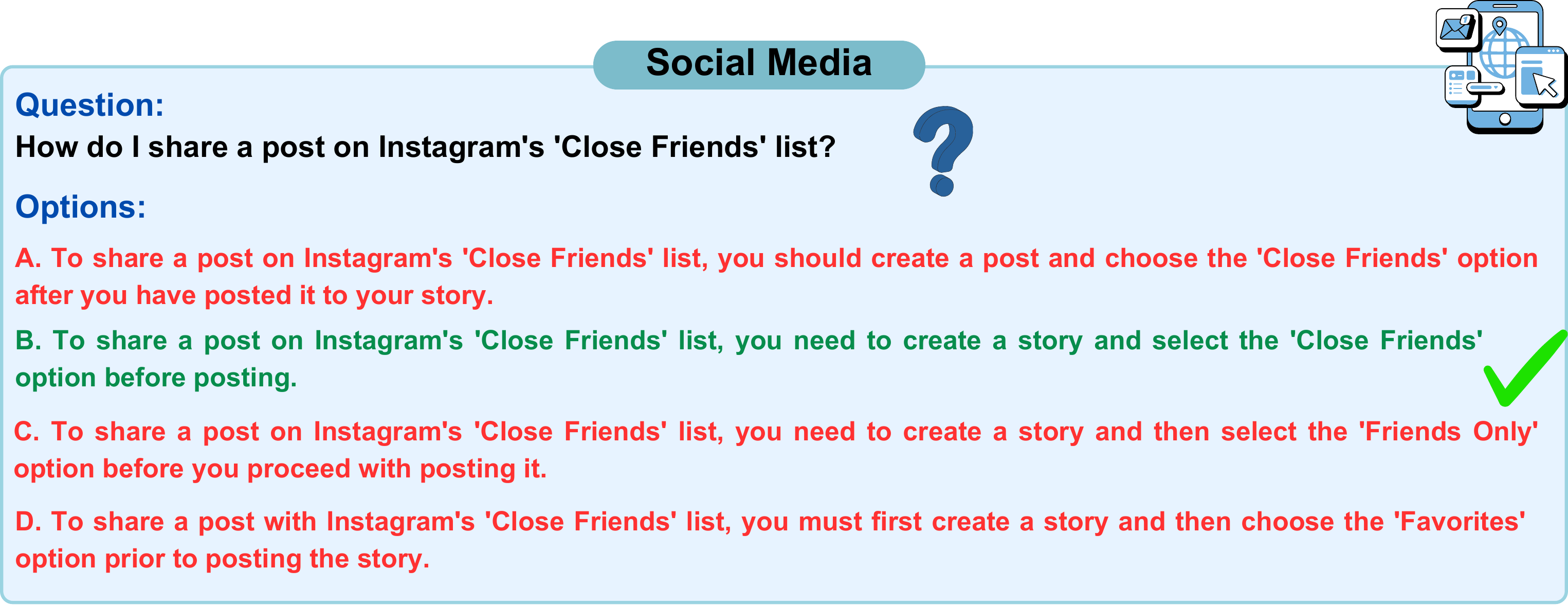}
  \caption{Example question from the {\em Social Media} field in Mobile-MMLU dataset.}
  \label{social_media_example1}
  \vspace{0.1in}
\end{figure}

\begin{figure}[t]
  \centering
    \includegraphics[width=1\linewidth]{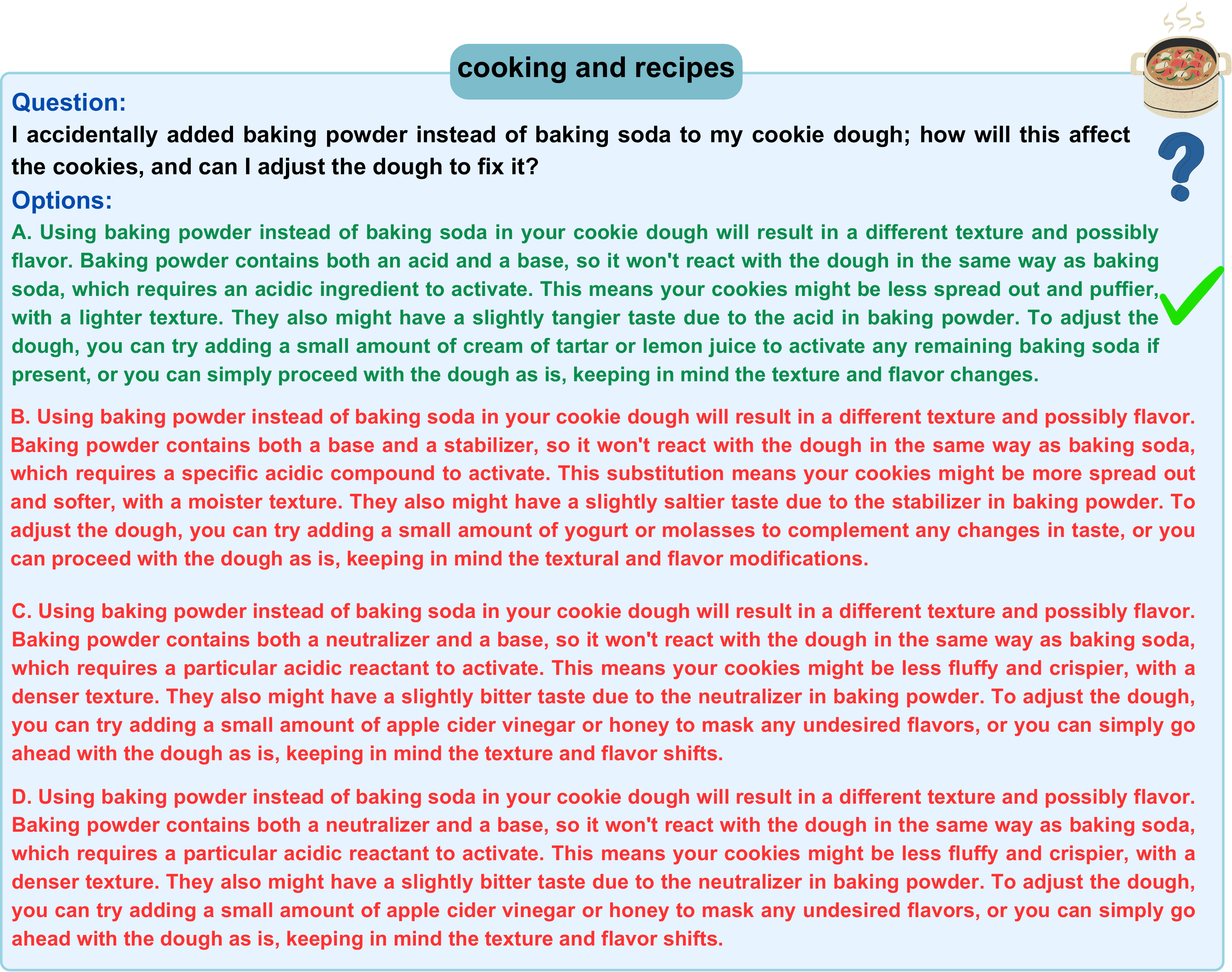}
  \caption{Example question from the {\em Cooking and Recipes} field in Mobile-MMLU dataset.}
  \label{cooking_and_recipe_example1}
  \vspace{0.1in}
\end{figure}

Currently, most existing multiple-choice question benchmarks share a common issue: they produce inconsistent performance results when evaluated repeatedly across various runs or different strong LLMs. To mitigate this variance, we design our dataset specifically for mobile scenarios and models, aiming for consistent results on server-side models as well, minimizing variance. Moreover, we seek to increase the difficulty level and reduce testing overhead by decreasing the number of samples. To achieve these goals, we implement two steps: (1) filtering out questions consistently answered correctly by two groups of mobile-level LLMs, and (2) removing questions where predictions differ among the strongest LLMs, including GPT-4o, Claude-3.5, and Gemini-2.0.
 
The resulting Mobile-MMLU-Pro retains the comprehensive coverage of the original dataset while being more suitable for resource-constrained environments. Table \ref{table:comparison} provides a detailed comparison between Mobile-MMLU and Mobile-MMLU-Pro, highlighting key differences in size and difficulty.

Specifically, Mobile-MMLU-Pro is a subset of Mobile-MMLU, which is created by evaluating Mobile-MMLU on two groups of models: a small-scale-model ensemble (Qwen-3B, LLaMA-3.2 3B, Gemma-2 2B, and Phi-3.5) and a moderate-scale-model ensemble (Qwen-7B, LLaMA-8B, and Gemma-9B) (dual-model evaluation approach). Drawing inspiration from recent work in rejection sampling for language models \cite{apple2024foundation, yuan2023scaling, yang2024qwen2}, we develop a two-phase evaluation framework for Mobile-MMLU-Pro that combines model-based filtering with selective rejection sampling.
A consistency constrain is further applied by removing questions where predictions differ among the strongest models of GPT-4o, Claude-3.5, and Gemini-2.0.

Our approach differs from previous rejection sampling methods in several key aspects. While Yuan et al.~\cite{yuan2023scaling} focus on using rejection sampling to identify correct reasoning paths for mathematical problems, and Apple \cite{apple2024foundation} employs a teacher committee (iTeC) for iterative refinement, our methodology specifically targets the identification of discriminative questions that can effectively differentiate model capabilities. This is achieved through a carefully orchestrated combination of multiple model-based evaluation and selective rejection sampling. The resulting Mobile-MMLU-Pro benchmark demonstrates strong discriminative power and consistency across different model scales while maintaining focus on practical, mobile-relevant scenarios.

\begin{table}[h]
\centering
\resizebox{0.99\linewidth}{!}{
\begin{tabular}{>{\raggedright\arraybackslash}m{5cm}  m{4cm}  m{4cm}}
\toprule
\textbf{Aspect} & \textbf{Mobile-MMLU} & \textbf{Mobile-MMLU-Pro} \\ 
\hline
\multirow{2}{5cm}{\textbf{Data Quality}} & High Quality & High Quality \\
 & \text{(General mobile-centric dataset)} & \text{(Refined dataset)} \\ 
\hline
\multirow{2}{5cm}{\textbf{Difficulty}} & Easier for most models & More difficult for all models \\
 & \text{(Appropriate for general use)} & \text{(Targets challenging queries)} \\ 
\hline
\multirow{2}{5cm}{\textbf{Question Types}} & Broad range of question types & Focus on harder, more complex questions \\
 & \text{(General mobile knowledge)} & \text{(Advanced difficulty)} \\ 
\hline
\multirow{2}{5cm}{\textbf{Size}} & Larger in size & Smaller in size \\
 & \text{(16,186)} & \text{(9,497)} \\ 
\hline
\multirow{2}{5cm}{\textbf{Scenario Used}} & Suitable for general use & Ideal for users seeking small, challenging datasets \\
 & \text{(Wide applicability)} & \text{(Consistent model performance)} \\
\bottomrule
\end{tabular}
}
\caption{Comparison of Mobile-MMLU and Mobile-MMLU-Pro datasets.}
\label{table:comparison}
\vspace{-0.15in}
\end{table}

\begin{figure}[t]
    \centering
    \includegraphics[width=0.95\textwidth]{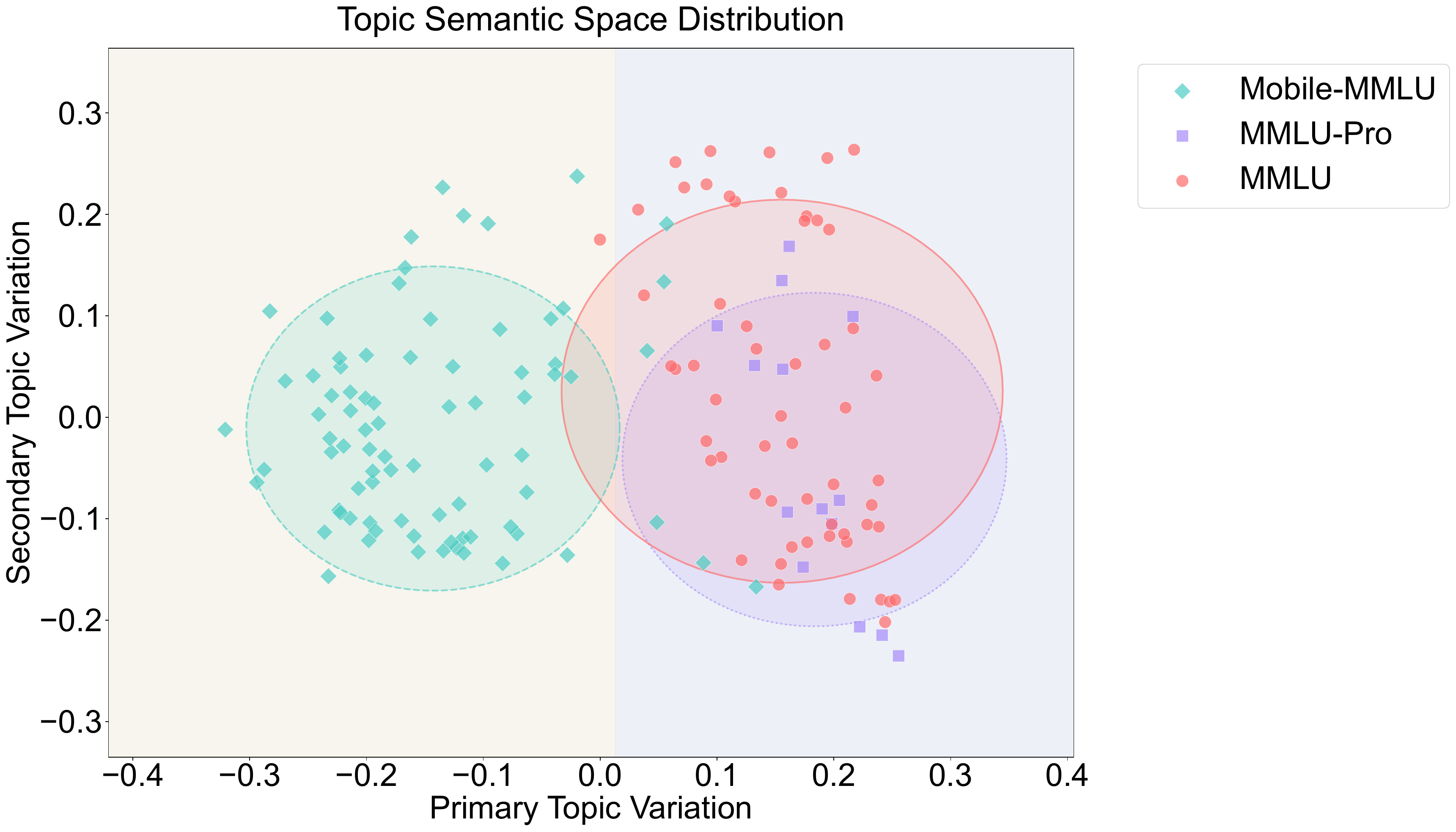}
    \caption{Topic distribution across Mobile-MMLU, MMLU, and MMLU-Pro benchmarks.}
    \label{fig:topic}
    \vspace{0.1in}
\end{figure}

\section{ Data Statistics and Distribution Analysis} \label{sec:data-stat}
Our analysis shows a significant gap in the existing general purpose benchmarks for evaluating mobile-specific LLM capabilities. While benchmarks like MMLU \cite{hendrycks2020measuring} and MMLU-Pro \cite{wang2024mmlu} are good for assessing broad knowledge and general abilities, they include numerous topics rarely encountered in mobile contexts, such as advanced theoretical concepts or extensive coding tasks. They underrepresent everyday mobile scenarios like quick recipe lookups, travel recommendations, or context-aware assistance. 

This misalignment between benchmark content and actual mobile use cases can lead to suboptimal model selection and evaluation for mobile deployments. Figure \ref{fig:topic}, demonstrates the topic distribution across Mobile-MMLU, MMLU, and MMLU-Pro benchmarks. We use ``all-mpnet-base-v2'' \cite{song2020mpnet} model from Sentence-Transformer \cite{reimers-2019-sentence-bert} to get get the sentence embedding for each of the questions from all three benchmark. We represent the topic using the average of all the question embedding in that topic and then use PCA~\cite{mackiewicz1993principal} to get two dimensional representation of the topics.
From the scatter plot, we can observe that Mobile-MMLU topics occupy a distinct semantic space compared to the topics of MMLU and MMLU-Pro benchmark. This clear separation in the topic distribution highlights Mobile-MMLU's unique focus on practical, mobile-relevant scenarios, complementing existing benchmarks rather than overlapping with them. The distinct clustering pattern validates Mobile-MMLU's contribution as a specialized benchmark tailored for evaluating mobile-oriented language models.

\begin{figure}[t]
    \centering
    \begin{subfigure}[b]{0.48\textwidth}
        \centering
        \includegraphics[width=\textwidth]{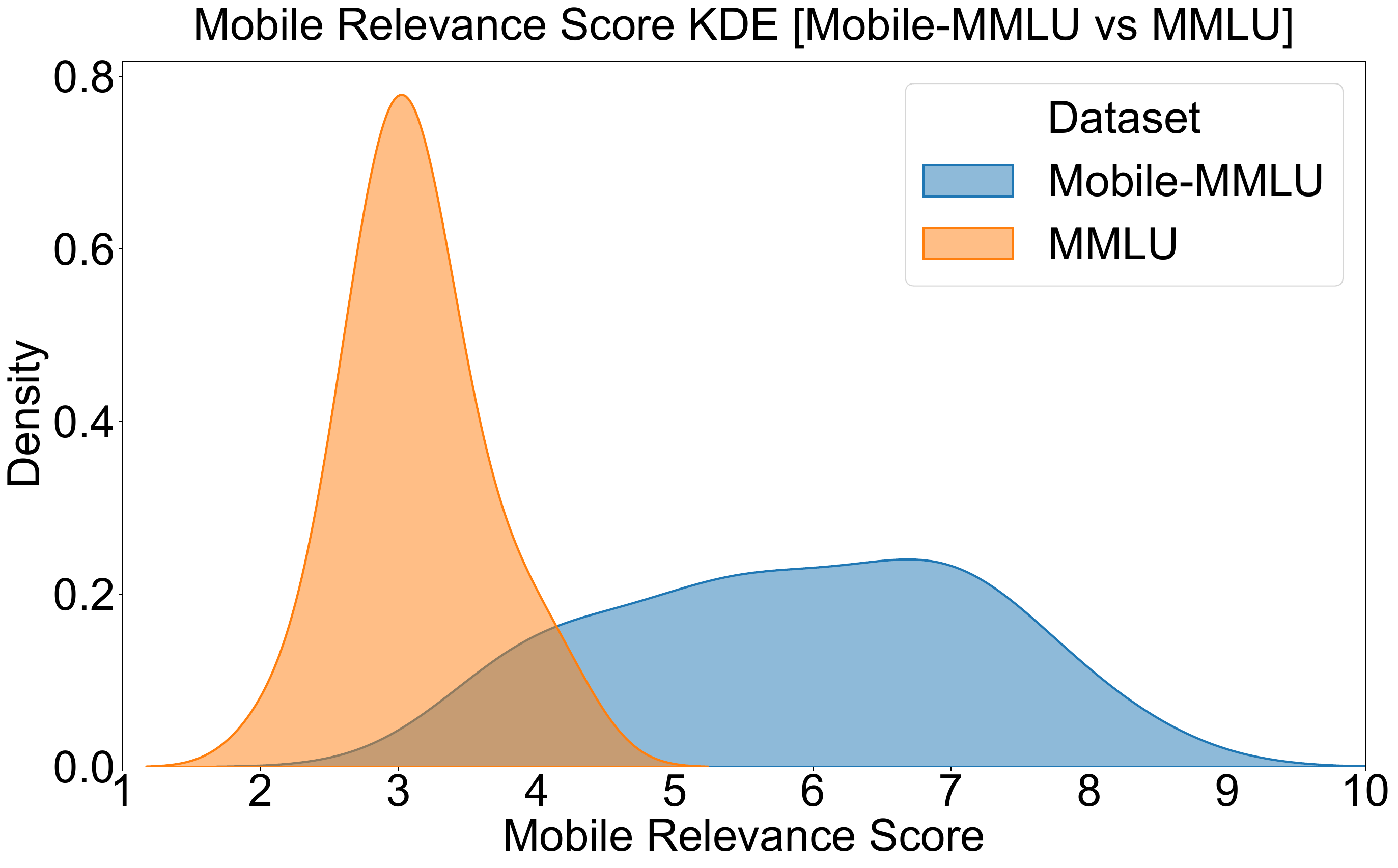}
        \caption{Mobile-MMLU vs MMLU}
        \label{fig:mscore_mmlu}
    \end{subfigure}
    \hfill
    \begin{subfigure}[b]{0.48\textwidth}
        \centering
        \includegraphics[width=\textwidth]{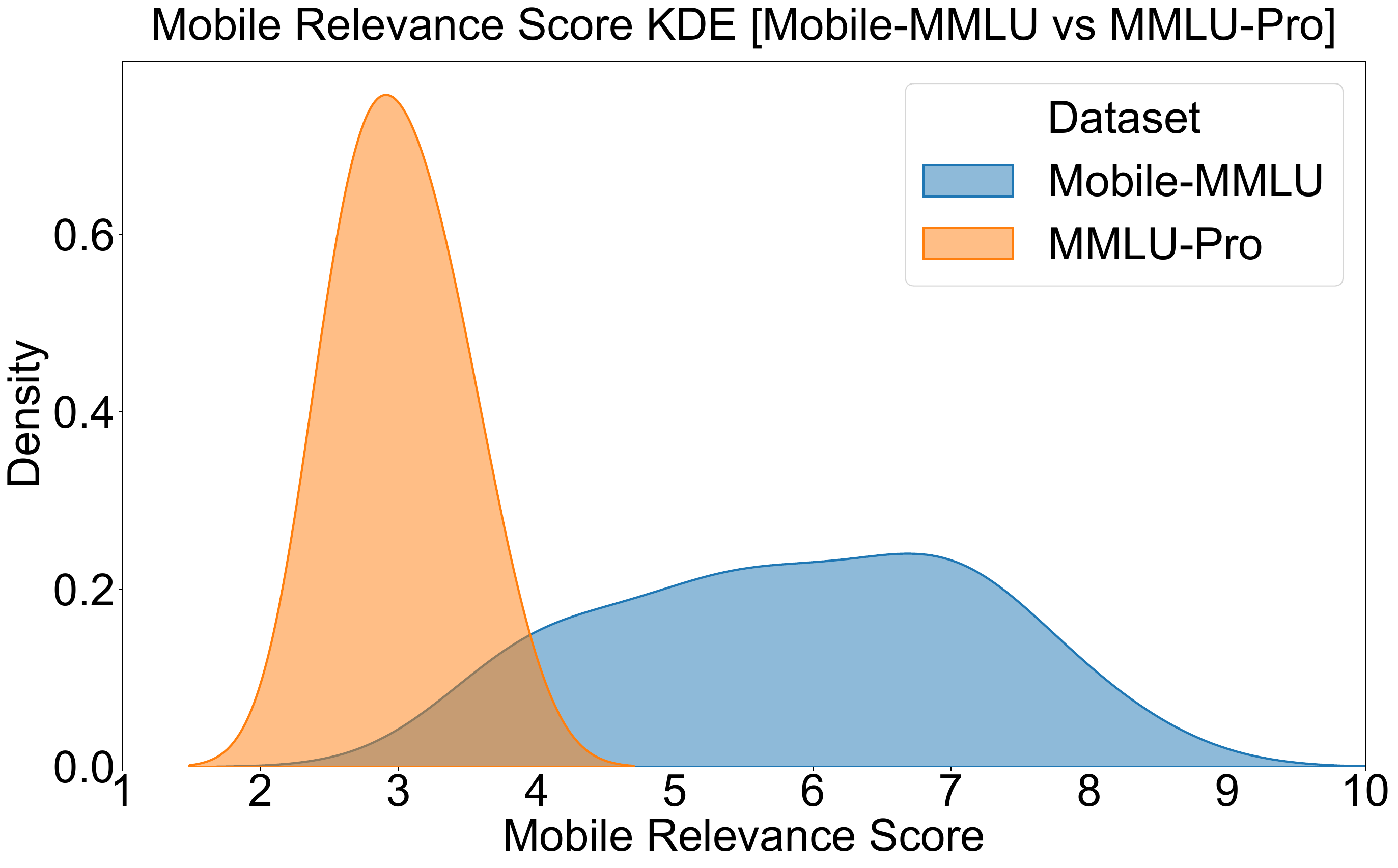}
        \caption{Mobile-MMLU vs MMLU-Pro}
        \label{fig:mscore_mmlupro}
    \end{subfigure}
    \caption{Distribution of MRScore comparing Mobile-MMLU against general benchmarks.}
    \label{fig:mscore_comparison}
    \vspace{0.1in}
\end{figure}

In order to further validate our hypothesis, we use GPT-4o~\cite{achiam2023gpt} as a judge \cite{zheng2023judging} to evaluate the Mobile Relevance Score (MRScore) of different questions in our benchmark. We then aggregate these mobile relevance scores of the questions at the topic level. 
Given a system prompt $S$ defining mobile expertise and evaluation guidelines, and a user prompt $C$ specifying evaluation criteria (practical value, mobile-friendliness, usage patterns), we define the MRScore for a question $q$ as:
\begin{equation}
   \begin{split}
   \text{MRScore}(q) = f_{\text{LLM}}(q|S,C) 
   \end{split}
   \label{eq:mrscore}
\end{equation}
This process can be decomposed implicitly through {\em $\text{round}(\lambda_1 \cdot P(q) + \lambda_2 \cdot M(q) + \lambda_3 \cdot U(q)) \in [1,10]$},
where $P(q)$, $M(q)$, and $U(q)$ represent practical value, mobile-friendliness, and usage pattern scores respectively, with $\lambda_1, \lambda_2, \lambda_3$ as implicit weights. For a topic $T$ containing $n$ questions, the aggregate score can be formulated as:
\begin{equation}
    \text{TopicMRScore}(T) = \frac{1}{n}\sum_{q \in T} \text{MRScore}(q)
    \label{eq:topic_mrscore}
\end{equation}
where each $\text{MRScore}(q)$ is conditioned on the same system and user prompts.
Figures \ref{fig:mscore_mmlu} and \ref{fig:mscore_mmlupro} show the distribution of MRScores across different benchmarks. 
The analysis reveals that questions in Mobile-MMLU consistently achieve higher MRScores compared to both MMLU and MMLU-Pro. The distributions show clear separation, with Mobile-MMLU questions predominantly scoring between 5 to 9, while traditional benchmarks cluster around 2 to 4. This quantitative validation confirms that our benchmark effectively captures mobile-specific use cases and scenarios.

\begin{figure}[t]
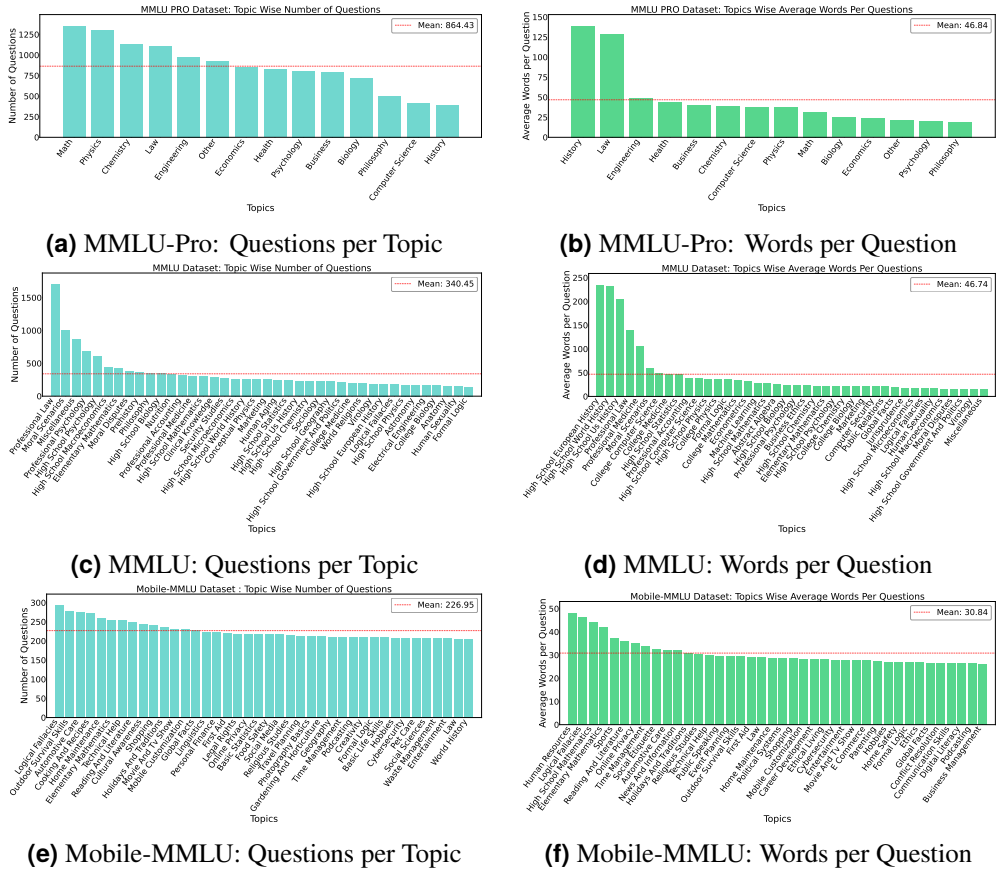

    \centering
    \begin{subfigure}[b]{0.48\textwidth}
        \centering
        \includegraphics[width=\textwidth]{images/mmlu_pro_questions_top14.pdf}
        \caption{MMLU-Pro: Questions per Topic}
        \label{fig:mmlu_pro_q}
    \end{subfigure}
    \hfill
    \begin{subfigure}[b]{0.48\textwidth}
        \centering
        \includegraphics[width=\textwidth]{images/mmlu_pro_words_top14.pdf}
        \caption{MMLU-Pro: Words per Question}
        \label{fig:mmlu_pro_w}
    \end{subfigure}
    
    \begin{subfigure}[b]{0.48\textwidth}
        \centering
        \includegraphics[width=\textwidth]{images/mmlu_questions_top40.pdf}
        \caption{MMLU: Questions per Topic}
        \label{fig:mmlu_q}
    \end{subfigure}
    \hfill
    \begin{subfigure}[b]{0.48\textwidth}
        \centering
        \includegraphics[width=\textwidth]{images/mmlu_words_top40.pdf}
        \caption{MMLU: Words per Question}
        \label{fig:mmlu_w}
    \end{subfigure}
    
    \begin{subfigure}[b]{0.48\textwidth}
        \centering
        \includegraphics[width=\textwidth]{images/mobile_mmlu_questions_top40.pdf}
        \caption{Mobile-MMLU: Questions per Topic}
        \label{fig:mobile_q}
    \end{subfigure}
    \hfill
    \begin{subfigure}[b]{0.48\textwidth}
        \centering
        \includegraphics[width=\textwidth]{images/mobile_mmlu_words_top40.pdf}
        \caption{Mobile-MMLU: Words per Question}
        \label{fig:mobile_w}
    \end{subfigure}
    \caption{Distribution of questions and average words per question across topics in MMLU-Pro, MMLU, and Mobile-MMLU datasets (showing top 40 topics). The red dashed line indicates the mean value for each metric.}
    \label{fig:questions_word_distri}
    \vspace{0.1in}
\end{figure}

 Our Mobile-MMLU consists of 80 topics, 16,186 questions, curated with the process discussed in the previous sections to evaluate small language models for the mobile specific use cases. Each topic includes multiple-choice questions designed to test both fundamental knowledge and real-world applications. Figure \ref{fig:questions_word_distri} shows the distribution of questions across top 40 topics for mobile-mmlu and mmlu dataset and all 14 topics for mmlu-pro, demonstrating the breadth and depth of our benchmark's coverage. Our questions focus on practical mobile usage scenarios and everyday tasks, ranging from ``Cooking And Recipes'' to ``Digital Literacy'' and ``Travel-Planning'', while also covering essential knowledge areas. Compared to traditional benchmarks like MMLU and MMLU-Pro, our questions are deliberately crafted to be more mobile-friendly, better reflecting real-world mobile interactions and information-seeking patterns. 

Figure \ref{fig:questions_word_distri} also shows that our Mobile-MMLU and Mobile-MMLU-Pro differ significantly from both MMLU and MMLU-Pro. MMLU-Pro concentrates heavily on specialized academic fields with Math (1,350 questions), Physics (1,300 questions), and Chemistry (1,150 questions) dominating the distribution, and MMLU emphasizes traditional educational subjects like Professional Law (1,700 questions) and Moral Scenarios (1,000 questions), in contrast, Mobile-MMLU maintains a more balanced distribution with practical topics such as Logical Fallacies (290 questions), Survival Skills (275 questions), and Automotive Care (270 questions) that are more relevant to everyday mobile information needs. The question length analysis further highlights this distinction, Mobile-MMLU questions average around 30.84 words per question, with even the most detailed topics like Human Resources not exceeding 48 words, making them more suitable for mobile interfaces. This is in stark contrast to MMLU and MMLU-Pro, which average 46.74 and 46.84 words per question, respectively, with some topics like European History in MMLU containing questions averaging over 230 words in length.

\begin{table}[h]
        \resizebox{\textwidth}{!}{ 
        \begin{tabular}{@{} l | c c | c c @{}}
            \toprule
            \textbf{Model} & 
            \makecell{\textbf{MMLU} \\ \textbf{(0-shot)}} & 
            \makecell{\textbf{MMLU-Pro} \\ \textbf{(0-shot)}} & 
            \makecell{\textbf{Mobile-MMLU} \\ \textbf{(0-shot)}} &  
            \makecell{\textbf{Mobile-MMLU-Pro} \\ \textbf{(0-shot)}} \\
            \midrule
            \multicolumn{5}{c}{\textbf{Mobile-Friendly Models}} \\ 
            \midrule
            Nemotron-Mini-4B-Instruct & 56.8 & 18.1 & 35.1 & 30.8\\
            Llama-3.2-1B-Instruct & 45.9 & 7.5 & 34.5 & 31.1\\
            Gemma-2-2b-it  & 56.8 & 17.2 & 38.9 & 31.2 \\
            Falcon3-3B-Instruct & 55.8 & 22.3 & 42.6 & 37.2\\
            Granite-3.1-3b-a800m-instruct & 50.5 & 12.8 & 43.6 & 39.4\\
            Llama-3.2-3B-Instruct  & 60.3 & 24.4 & 50.2 & 42.0\\
            Granite-3.1-2b-instruct & 54.3 & 20.2 & 48.1 & 42.2\\
            Qwen2.5-1.5B-Instruct & 60.1 & 19.9 & 49.7 & 43.0\\
            Exaone-3.5-2.4B-Instruct & 58.1 & 25.3 & 53.7 & 47.7\\
            Phi-3.5-mini-instruct  & 68.7 & 32.9 & 63.7 & 54.8\\
            Qwen2.5-3B-Instruct  & 65.4 & 25.0 & 68.1 & 60.6\\
            \midrule
            \multicolumn{5}{c}{\textbf{General-Purpose Models}} \\
            \midrule
            Olmo-2-1124-7B-Instruct & 59.3 & 18.6 & 49.6 & 42.9\\
            Falcon3-7B-Instruct & 68.0 & 34.3 & 52.4 & 46.8\\
            Falcon3-10B-Instruct & 71.6 & 38.1 & 54.3 & 49.1\\
            Yi-1.5-6B-Chat & 61.8 & 24.4 & 60.5 & 54.7\\
            Llama-3.1-8B-Instruct  & 67.9 & 30.7 & 66.9 & 57.1\\
            Granite-3.1-8b-instruct & 64.5 & 28.2 & 60.8 & 57.7\\
            Internlm2\_5-7b-chat & 67.7 & 30.4 & 64.3 & 58.6\\
            Ministral-8B-Instruct-2410 & 64.0 & 25.4 & 71.5 & 63.6\\
            Yi-1.5-9B-Chat & 68.3 & 33.1 & 72.7 & 67.7\\
            Qwen2.5-7B-Instruct  & 71.7 & 36.5 & 74.9 & 68.4\\
            Gemma-2-9b-it  & 71.8 & 31.9 & 75.0 & 69.1\\
            \bottomrule
        \end{tabular}
        }
    \caption{Comparison of performance on Mobile-MMLU and Mobile-MMLU-Pro with prior MMLU and MMLU-Pro benchmarks for different sizes of models.}
    \label{bench_res_table}
    \vspace{0.1in}
\end{table}

\section{Experiments}

\subsection{Setup}

\noindent{\bf Models:} To comprehensively evaluate model performance on Mobile-MMLU and Mobile-MMLU-Pro, we select a diverse set of state-of-the-art language models ranging from 1B to 9B parameters. Our model suite includes Gemma-2-9B-it \cite{team2024gemma}, Qwen2.5-7B-instruct \cite{yang2024qwen2}, Llama-3.1-8B-instruct \cite{dubey2024llama}, Qwen2.5-3B-instruct \cite{yang2024qwen2}, Phi-3.5-mini-instruct \cite{abdin2024phi}, Llama-3.2-3B-instruct \cite{meta2024llama}, Gemma-2-2B-it \cite{team2024gemma}, Ministral-8B-instruct, Qwen2.5-1.5B-instruct \cite{yang2024qwen2} and Llama-3.2-1B-instruct \cite{meta2024llama}. This selection encompasses models of varying architectures and parameter counts, enabling a thorough analysis of the relationship between model size and performance on mobile-oriented tasks.

\noindent{\bf Evaluation:} For our evaluation framework, we use \textbf{lm-eval-harness} \cite{eval-harness} to assess the performance of the models. Given that Mobile-MMLU and Mobile-MMLU-Pro consist entirely of multiple-choice questions, we focus on accuracy as our primary evaluation metric. This approach allows for objective comparison across models while maintaining consistency with existing benchmarking practices in the field.

\subsection{Main Results} \label{main}
\noindent{\bf Overall Performance Comparison}: Our evaluation reveals several significant patterns across different model scales and architectures. Table \ref{bench_res_table} presents the zero-shot performance of both mobile-friendly and general-purpose (slightly larger size) models across MMLU, MMLU-Pro, Mobile-MMLU, and Mobile-MMLU-Pro benchmarks. 
On our benchmark, the performance gap between the strongest and weakest models is larger compared to other benchmarks, highlighting the greater discriminative capability of our dataset, high-performing models achieve better scores, while weaker models perform worse. For instance, the lowest-performing model, Nemotron-Mini-4B-Instruct, scores 35.1\% on our dataset but achieves 56.8\% on MMLU. Conversely, the highest-performing model, Qwen2.5-3B-Instruct, obtains a higher score of 68.1\% on our benchmark, surpassing its 65.4\% score on MMLU. We emphasize that a strong benchmark dataset should not aim merely to lower model performance scores, rather, it should accentuate distinctions between models, clearly highlighting significant performance gaps and enabling better differentiation of their capabilities.
In addition, our results demonstrate that strong performance on traditional benchmarks does not necessarily translate to superior performance on mobile-specific tasks. A notable example is Phi-3.5-mini-instruct, which achieves impressive scores on MMLU (68.7\%) but shows relatively lower performance on Mobile-MMLU (63.7\%). Conversely, Qwen2.5-3B-Instruct, despite its modest performance on MMLU (65.4\%), excels on Mobile-MMLU (68.1\%), even outperforming some larger 8B parameter models.

\begin{table}[t]
    \centering
    \renewcommand{\arraystretch}{1.2}
    \resizebox{\textwidth}{!}{ 
    \begin{tabular}{lc|cccc|cccc}
        \toprule
        Model & Ori. & A & B & C & D & {\em R} \#1 & {\em R} \#2 & {\em R} \#3 & {\em R} \#4 \\
        \midrule
        Phi-3.5-mini-instruct & 63.7 & 45.77 & 70.67 & 76.99 & 87.66 & 65.61 & 65.89 & 65.05 & 65.32 \\
        Qwen2.5-3B-Instruct & 68.1 & 53.74 & 79.04 & 74.30 & 83.29 & 70.22 & 69.85 & 70.36 & 69.99 \\
        Llama-3.2-3B-Instruct & 50.2 & 65.17 & 61.74 & 57.12 & 56.54 & 55.82 & 54.81 & 54.64 & 54.97 \\
        Llama-3.1-8B-Instruct & 66.9 & 50.78 & 88.45 & 78.94 & 75.45 & 68.75 & 68.87 & 67.45 & 67.83 \\
        Qwen2.5-7B-Instruct & 74.9 & 54.82 & 85.39 & 86.23 & 88.72 & 75.91 & 75.26 & 73.43 & 74.19 \\
        gemma-2-9b-it & 75.0 & 90.63 & 70.30 & 68.25 & 80.46 & 76.35 & 76.72 & 75.39 & 75.21 \\
        gemma-2-2b-it & 38.9 & 67.67 & 39.40 & 41.63 & 37.37 & 42.11 & 41.74 & 40.85 & 41.43 \\
        \bottomrule
    \end{tabular}
    }
    \caption{Ablation of {\em Selection Order Bias} using different models on Mobile-MMLU.}
    \label{tab:model_performance_order_full}
    \vspace{-0.1in}
\end{table}

\begin{table}[t]
    \centering
    \renewcommand{\arraystretch}{1.2}
    \resizebox{\textwidth}{!}{ 
    \begin{tabular}{lc|cccc|cccc}
        \toprule
        Model & Ori. & A & B & C & D & {\em R} \#1 & {\em R} \#2 & {\em R} \#3 & {\em R} \#4 \\
        \midrule
        Phi-3.5-mini-instruct & 54.8 & 41.19 & 62.78 & 66.75 & 73.94 & 57.53 & 52.74 & 55.89 & 57.11 \\
        Qwen2.5-3B-Instruct & 60.6 & 53.21 & 71.22 & 68.36 & 71.28 & 63.16 & 66.12 & 65.39 & 65.62 \\
        Llama-3.2-3B-Instruct & 42.0 & 54.36 & 49.96 & 50.54 & 48.88 & 47.12 & 46.38 & 44.72 & 42.69 \\
        Llama-3.1-8B-Instruct & 57.1 & 48.89 & 75.18 & 66.21 & 63.45 & 54.81 & 56.11 & 56.21 & 53.87 \\
        Qwen2.5-7B-Instruct & 68.4 & 54.52 & 77.55 & 76.98 & 79.14 & 71.24 & 69.29 & 69.15 & 70.37 \\
        gemma-2-9b-it & 69.1 & 85.14 & 65.32 & 63.11 & 73.12 & 70.25 & 68.99 & 70.72 & 70.15 \\
        gemma-2-2b-it & 31.2 & 51.65 & 34.11 & 36.77 & 29.91 & 32.21 & 33.50 & 30.76 & 31.51 \\
        \bottomrule
    \end{tabular}
    }
    \caption{Ablation of {\em Selection Order Bias} using different models on Mobile-MMLU-Pro.}
    \label{tab:model_performance_order_pro}
\end{table}

\subsection{Analysis}

\noindent{\bf Performance Distribution Analysis}: One of the most striking observations is the wide performance variance across models on Mobile-MMLU compared to other benchmarks. While the performance spread on MMLU ranges from 45.9\% to 71.8\%, and MMLU-Pro from 7.5\% to 36.5\%, Mobile-MMLU exhibits a wider relative range from 34.5\% to 75.0\%. This increased variance is particularly pronounced among smaller models (1-3B parameters), providing valuable insights for mobile deployment scenarios where model size constraints are critical. Furthermore, it can be seen that model size alone does not determine performance. For instance, among similarly-sized models in the 3B parameter range, we observe substantial performance differences: Qwen2.5-3B-Instruct achieves 68.1\% accuracy on Mobile-MMLU, while Llama-3.2-3B-Instruct scores 50.2\%, despite both models having comparable parameter counts.

\noindent{\bf LLM Selection Order Bias}: To investigate how the position of the correct answer affects model predictions, we examine the following settings:
\begin{itemize}
    \item {\em Our Original Dataset Construction Strategy}: Initially, the correct options (A/B/C/D) are randomly assigned and then filtered, resulting in a slight non-uniform distribution in the final dataset but the results are still close and stable to the full random-order performance (as shown in the first group of Tables~\ref{tab:model_performance_order_full} and~\ref{tab:model_performance_order_pro}).
    \item {\em Systematic Placement}: The correct answers are systematically rotated across A/B/C/D, while the incorrect options are randomly assigned (as shown in the middle group of Tables~\ref{tab:model_performance_order_full} and~\ref{tab:model_performance_order_pro}).
    \item {\em Fully Randomized Distribution}: Both correct and incorrect options are randomly distributed (as shown in the last group of Tables~\ref{tab:model_performance_order_full} and~\ref{tab:model_performance_order_pro}).
\end{itemize}

\begin{figure}[t]
  \centering
  \includegraphics[width=0.95\linewidth]{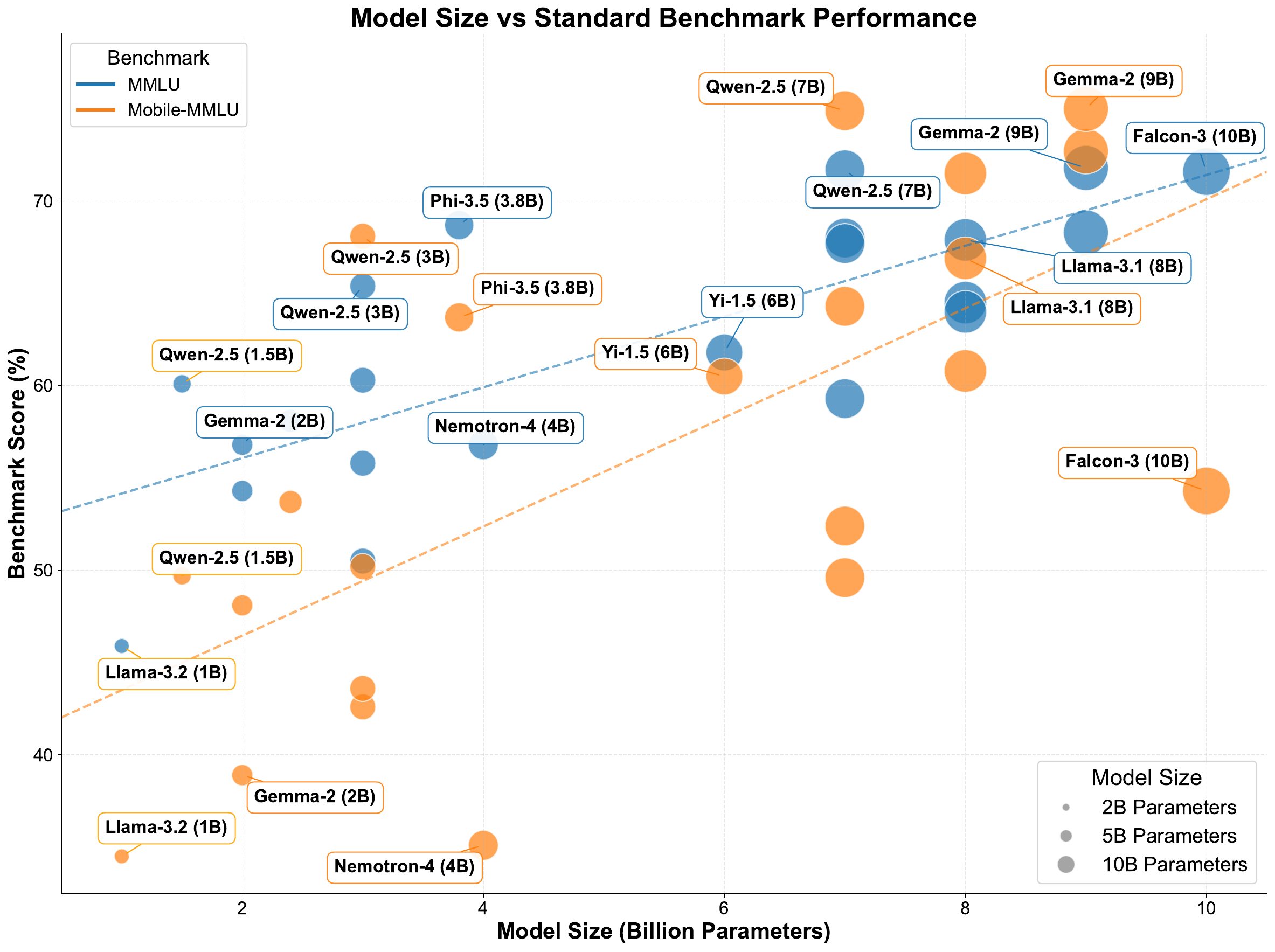}
  \caption{Model Size vs. Performance of Mobile-MMLU and MMLU.}
  \label{fig:standard_benchmarks_comparison}
\end{figure}

\begin{figure}[t]
  \centering
  \includegraphics[width=0.95\linewidth]{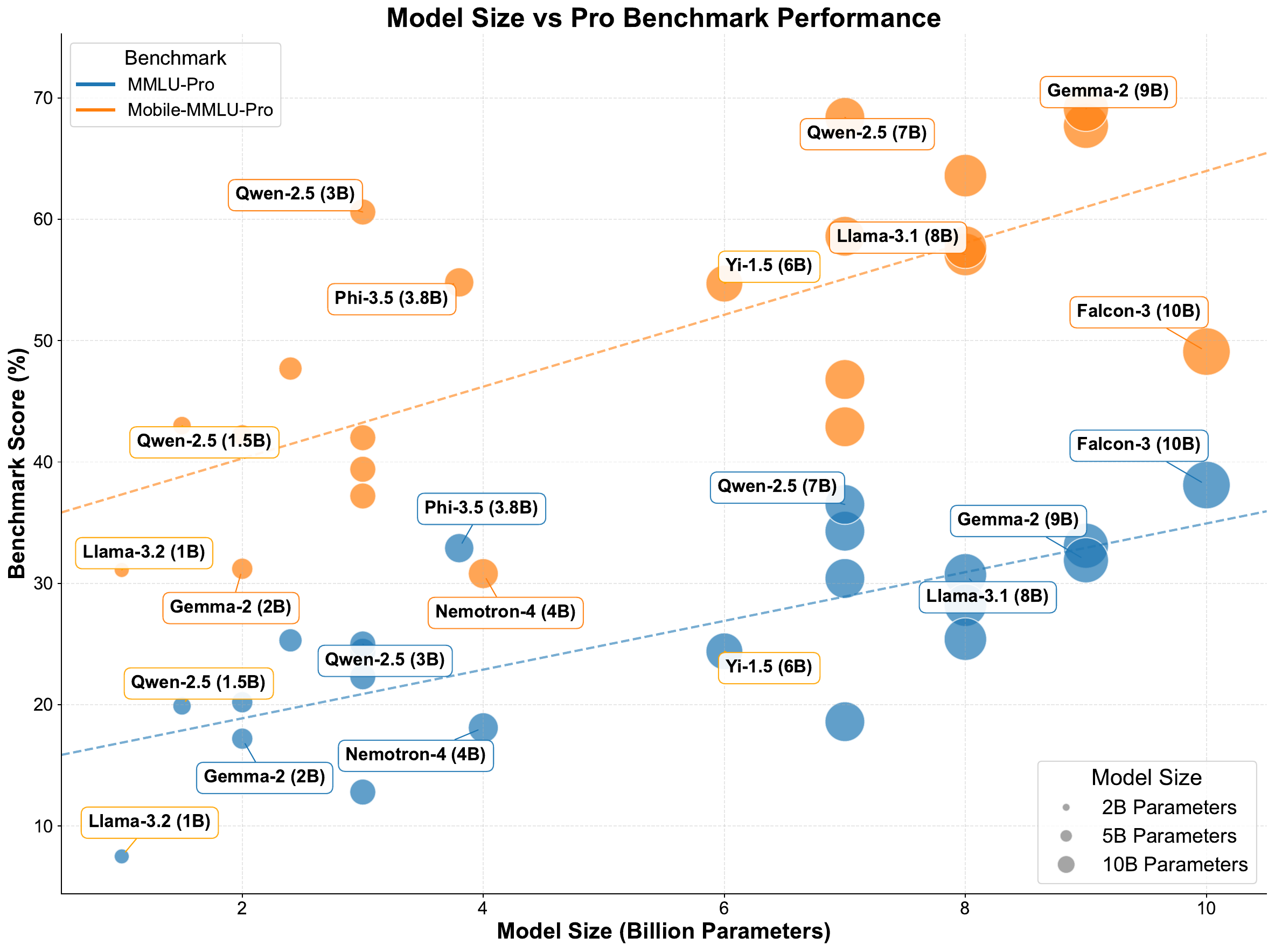}
  \caption{Model Size vs. Performance of Mobile-MMLU-Pro and MMLU-Pro.}
  \label{fig:pro_benchmarks_comparison}
\end{figure} 

The {\em Ori.} columns in Tables~\ref{tab:model_performance_order_full} and~\ref{tab:model_performance_order_pro} present the results on our {\bf Mobile-MMLU} and {\bf Pro} benchmarks, respectively. Our findings show strong consistency with the random-order results, exhibiting minimal variance. Thus, we refer to our dataset as {\em order-invariant}, reflecting its stable properties within these benchmarks.
Our results in middle groups of these two tables also indicate that small LLMs are highly sensitive to the order of answer choices, with performance variance reaching to more than 10\%. This highlights the importance of fair dataset construction, particularly through systematic balancing. However, the most robust solution is to adopt the open-llm-leaderboard~\cite{myrzakhan2024open} scheme, converting multiple-choice tasks into open-ended generation tasks, which eliminates this issue entirely. We plan to conduct further research in this area.

\subsection{Visualization of Model Size vs. Benchmark Performance}

Figures~\ref{fig:standard_benchmarks_comparison} and~\ref{fig:pro_benchmarks_comparison} show the relationship between model parameter and performance across our standard and pro version of benchmarks. For the standard benchmark comparison in Figure~\ref{fig:standard_benchmarks_comparison}, MMLU (blue dotted line) consistently shows higher performance than Mobile-MMLU (orange dotted line) across most model sizes, with scores 5$\sim$10\% higher for comparable models. Also, as we have mentioned in Section~\ref{main}, model performance on MMLU is more concentrated and lacks distinction. In the pro version comparison of Figure~\ref{fig:pro_benchmarks_comparison}, Mobile-MMLU-Pro scores are higher than MMLU-Pro for the same models, while model performance on MMLU-Pro is still more intensive than models on our Mobile-MMLU-Pro. Both benchmark families show a positive correlation between model size and performance, but with notably different absolute scores. Some smaller models like Qwen2.5-3B-Instruct achieve competitive performance on both benchmark variants, suggesting that architectural improvements can sometimes compensate for smaller parameter counts.

\section{Conclusion}

This paper introduced Mobile-MMLU and Mobile-MMLU-Pro, a novel benchmark family designed to evaluate language models in mobile contexts, addressing a critical gap in assessing mobile-optimized language models. Through careful curation of 80 diverse topics with 16,186, our Mobile-MMLU focuses on practical, mobile-relevant scenarios that better reflect real-world mobile interactions. The complementary Mobile-MMLU-Pro benchmark, created through our rigorous multi-model-consistency based rejection sampling approach, provides a more challenging and consistent evaluation set while maintaining focus on mobile-specific use cases. Both of the two benchmarks enjoy {\em mobile-centric} and {\em order-invariant} properties. Our comprehensive analysis demonstrates that Mobile-MMLU and Mobile-MMLU-Pro occupy a distinct semantic space compared to traditional benchmarks like MMLU and MMLU-Pro, with consistently higher Mobile Relevance Scores.

Our evaluation results across various model sizes (1B-9B parameters) reveal critical insights about mobile language model capabilities. The wider performance variance observed in Mobile-MMLU, particularly among smaller models, highlights its effectiveness in differentiating model capabilities under mobile constraints. These findings demonstrate that strong performance on traditional benchmarks does not necessarily translate to superior performance on mobile-specific tasks, underscoring the importance of specialized evaluation frameworks for mobile applications. As mobile AI continues to evolve, Mobile-MMLU family provides a foundation for developing and assessing more efficient, capable, and user-centric mobile language models.
\section{Limitations and Ethical Statements}

\noindent{\bf Time-sensitive Questions in the Dataset.}
Upon manual verification, our dataset contains approximately 3.5\% of time-sensitive questions. These questions are included because they reflect the types of queries people commonly ask on their mobile phones, such as ``What is the best application?'', ``What is the population of a city?'', or ``How do I post a photo?'' Incorporating these questions is essential to creating a relevant mobile benchmark, as they represent practical, real-world use cases. While the ground truth answers for these questions are currently accurate, we acknowledge that they may become outdated over time, for instance, in 10 years, the answers might no longer be valid. To address this limitation, our benchmark will be regularly updated to maintain its relevance. 
We also introduce Mobile-MMLU-Pro, a subset of our dataset that excludes simple, sensitive, and inconsistent questions. This version offers a comparable size to MMLU-Pro and provides a more stable and consistent benchmark for evaluating model performance without the complications introduced by time-sensitive or location-sensitive queries.

\noindent{\bf Multi-choice to Open-ended Questions.}
Similar to many other datasets, our benchmark primarily focuses on multi-choice questions, potentially oversimplifying the complexity of real-world language tasks encountered by mobile users. This format may not fully represent the diverse linguistic interactions and contextual complexities in mobile environments, such as conversational nuances, multi-turn dialogue, or dynamic contextual information. Moreover, the benchmark's language diversity and domain coverage may not fully represent the global and multilingual nature of mobile user populations, restricting its comprehensive generalizability for different countries and races.

\section*{Acknowledgments}

This research is supported by the MBZUAI-WIS Joint Program for AI Research and the Google Research award grant.

\bibliography{art.bib,ref.bib}

\clearpage

\newgeometry{left=3cm,bottom=1cm}
\appendix
\section*{Appendix}

\section{Examples from Our Mobile-MMLU Benchmark}

We provide examples from our Mobile-MMLU benchmark to showcase the diversity of questions in our dataset. Each example consists of a question from one of the 80 fields covered, along with multiple-choice answer options and the correct answer. Tables~\ref{example1}, \ref{example6}, and \ref{example7} present examples from three distinct fields: \texttt{Technical Help}, \texttt{Home Maintenance}, and \texttt{Ergonomics}.

\begin{table}[h]
\centering
\scalebox{0.90}{
\begin{tabular}[c]{@{}p{16.5cm}@{}}
\toprule
\textbf{Field Name:} \texttt{Technical\_help}\\
\textbf{Question:} \texttt{How can I connect my phone to a car's Bluetooth system?}\\
\textbf{Options:}\\
\texttt{A. \textcolor{red}{  To pair your phone with a car's Bluetooth system, start by switching on Bluetooth on your phone. Next, enter the car's media system, find the Bluetooth options, and select "New Device" or "Join Device." Follow the detailed instructions on the screen to conclude the pairing procedure.}}\\
\texttt{B. \textcolor{red}{ To synchronize your phone with a car's Bluetooth system, initially turn on Bluetooth on your phone. Afterwards, head to the car's control panel, locate the connectivity settings, and pick "Discover Device" or "Link New Device." Follow the comprehensive prompts to finalize the connection process.}}\\
\texttt{C. \textcolor{blue}{To connect your phone to a car's Bluetooth system, first enable Bluetooth on your phone. Then, go to the car's infotainment system, navigate to the Bluetooth settings, and select "Add Device" or "Pair New Device." Follow the on-screen instructions to complete the pairing process.}}\\
\texttt{D. \textcolor{red}{  To connect your phone to a car's Bluetooth system, first ensure Bluetooth is activated on your phone. Then, access the car's infotainment system and proceed to the wireless settings menu. Choose "Link Device" or "Connect New Device," and adhere to the displayed guidelines to finish the pairing process.}}\\
\textbf{Correct Answer:} \texttt{c}\\
 \bottomrule
\end{tabular}
}
\caption{Example Question from the \texttt{Technical Help} Field in the Mobile-MMLU Dataset.}
\label{example1}
\end{table}

\begin{table}[h]
\centering
\scalebox{0.90}{
\begin{tabular}[c]{@{}p{16.5cm}@{}}
\toprule
\textbf{Field Name:} \texttt{Home maintenance}\\
\textbf{Question:} \texttt{What should I check if my washing machine is making loud banging noises during the spin cycle?(1) Ensure the washing machine is on a stable and flat surface, make sure the load is distributed evenly inside the drum, inspect the drum for any objects that might have been left behind, and check for worn or damaged drum bearings or suspension springs. (2) Check if the washing machine is balanced and level, ensure the load is evenly distributed, inspect the drum for foreign objects, and check for worn or damaged shock absorbers or suspension rods. (3) Check if the washing machine is correctly balanced and perfectly aligned, ensure the laundry load is properly distributed across the drum, inspect the inner drum for any foreign materials, and check for worn or damaged springs or suspension belts. (4) Verify that the washing machine is level and not tilted, confirm that the laundry is evenly spread within the drum, thoroughly inspect the drum for any foreign objects, and check for worn or damaged vibration dampers or suspension springs.
Which of the statements given above are correct?}\\
\textbf{Options:}\\
\texttt{A. \textcolor{red}{ (2) and (4) }}\\
\texttt{B. \textcolor{blue}{(1), (2) and (4)  }}\\
\texttt{C. \textcolor{red}{(1) only }}\\
\texttt{D. \textcolor{red}{(2), (3) and (4) }}\\
\textbf{Correct Answer:} \texttt{B}\\
 \bottomrule
\end{tabular}
}
\caption{Example Question from the \texttt{Home Maintenance} Field in the Mobile-MMLU Dataset.}
\label{example6}
\end{table}

\begin{table}[h]
\centering
\scalebox{0.90}{
\begin{tabular}[c]{@{}p{16.5cm}@{}}
\toprule
\textbf{Field Name:} \texttt{Ergonomics}\\
\textbf{Question:} \texttt{What is the best way to hold a smartphone to reduce strain?}\\
\textbf{Options:}\\
\texttt{A. \textcolor{red}{Hold the smartphone at waist level, which causes significant neck strain, with a relaxed grip to reduce strain while maintaining a comfortable posture .}}\\
\texttt{B. \textcolor{red}{Hold the smartphone slightly above eye level, which is uncomfortable, with a firm grip. }}\\
\texttt{C. \textcolor{red}{ Hold the smartphone at chest level, which still requires neck tilting, with a relaxed grip to minimize strain.}}\\
\texttt{D. \textcolor{blue}{Hold the smartphone at eye level with a relaxed grip to reduce strain.}}\\
\textbf{Correct Answer:} \texttt{D}\\
 \bottomrule
\end{tabular}
}
\caption{Example Question from the Ergonomics Field in the Mobile-MMLU Dataset.}
\label{example7}
\end{table}

\section{Data Hierarchy}
Table \ref{tab:hierarchy} shows the complete hierarchy of our Mobile-MMLU dataset. It covers most of the common knowledge domains relevant to mobile users' daily information needs. The dataset is organized into 9 major categories: \texttt{Academic \& Learning, Business \& Career, Technology \& Digital, Health \& Safety, Lifestyle \& Personal, Home \& Family, Culture \& Society, and Environment, with an additional category of Miscellaneous}. These categories are further divided into 27 subcategories and 80 distinct topics, with a total population of 16,186 question-answer pairs.
The \texttt{Lifestyle \& Personal} category contains the highest number of topics (17) covering various aspects of personal growth and daily interests. The \texttt{Academic \& Learning} category follows with 15 topics that address fundamental educational areas from elementary mathematics to world history. This hierarchical structure ensures comprehensive coverage of the knowledge domains that mobile users frequently encounter and query, bridging the gap in the existing general purpose benchmarks.

\FloatBarrier 
\footnotesize
\begin{longtable}[htbp]{lllr}
\caption{Hierarchical Structure of Categories, Subcategories, and Topics with Population of Mobile-MMLU.\label{tab:hierarchy}} \\
\toprule
\textbf{Category} & \textbf{Subcategory} & \textbf{Topic} & \textbf{Population} \\ 
\midrule
\endfirsthead

\multicolumn{4}{c}{Table \thetable{} continued from the previous page} \\
\midrule
\textbf{Category} & \textbf{Subcategory} & \textbf{Topic} & \textbf{Population} \\ 
\midrule
\endhead

\midrule \multicolumn{4}{r}{Continued on next page} \\
\endfoot

\bottomrule
\endlastfoot

\rowcolor{acad} Academic \& Learning & Basic Mathematics & Elementary Mathematics & 254 \\
\rowcolor{acad} & & High School Mathematics & 200 \\
\rowcolor{acad} & & Basic Statistics & 219 \\
\rowcolor{acad} & Basic Sciences & Conceptual Physics & 194 \\
\rowcolor{acad} & & Science Fundamentals & 191 \\
\rowcolor{acad} & Critical Thinking & Formal Logic & 210 \\
\rowcolor{acad} & & Logical Fallacies & 293 \\
\rowcolor{acad} & Education & Education Techniques & 146 \\
\rowcolor{acad} & & Reading \& Literature & 248 \\
\rowcolor{acad} & & Writing Skills & 203 \\
\rowcolor{acad} & & Linguistics & 223 \\
\rowcolor{acad} & Social Sciences & Social Sciences & 207 \\
\rowcolor{acad} & & Political Systems & 193 \\
\rowcolor{acad} & & World History & 206 \\
\rowcolor{acad} & & Geography & 211 \\
\midrule

\rowcolor{busi} Business \& Career & Business Studies & Project Management & 176 \\
\rowcolor{busi} & & Human Resources & 144 \\
\rowcolor{busi} & & Business Management & 167 \\
\rowcolor{busi} & & Marketing \& Sales Strategies & 162 \\
\rowcolor{busi} & Personal Business & Personal Finance & 223 \\
\rowcolor{busi} & & E Commerce & 186 \\
\rowcolor{busi} & & Shopping & 241 \\
\rowcolor{busi} & & Accounting & 205 \\
\rowcolor{busi} & Communication & Communication Skills & 134 \\
\rowcolor{busi} & & Social Etiquette & 200 \\
\rowcolor{busi} & & Public Speaking & 158 \\
\midrule

\rowcolor{tech} Technology \& Digital & Digital Literacy & Digital Literacy & 198 \\
\rowcolor{tech} & & Technical Help & 254 \\
\rowcolor{tech} & & Mobile Customization & 230 \\
\rowcolor{tech} & Privacy \& Security & Cybersecurity & 208 \\
\rowcolor{tech} & & Online Privacy & 219 \\
\rowcolor{tech} & Social Media & Social Media & 217 \\
\rowcolor{tech} & & Digital Detox & 196 \\
\midrule

\rowcolor{heal} Health \& Safety & Health \& Wellness & Mental Health & 130 \\
\rowcolor{heal} & & Physical Fitness & 190 \\
\rowcolor{heal} & & Medical \& Health Knowledge & 183 \\
\rowcolor{heal} & & Ergonomics & 204 \\
\rowcolor{heal} & Everyday Safety & First Aid & 221 \\
\rowcolor{heal} & & Outdoor Survival Skills & 277 \\
\rowcolor{heal} & & Automotive Care & 275 \\
\midrule

\rowcolor{life} Lifestyle \& Personal & Daily Life Skills & Basic Life Skills & 209 \\
\rowcolor{life} & & Time Management & 211 \\
\rowcolor{life} & & Conflict Resolution & 152 \\
\rowcolor{life} & & Event Planning & 201 \\
\rowcolor{life} & Personal Growth & Creativity & 210 \\
\rowcolor{life} & & Emotional Intelligence & 133 \\
\rowcolor{life} & & Personal Branding & 186 \\
\rowcolor{life} & & Career Development & 166 \\
\rowcolor{life} & Lifestyle & Fashion \& Style & 200 \\
\rowcolor{life} & & Travel Planning & 214 \\
\rowcolor{life} & & Sports & 188 \\
\rowcolor{life} & & Gardening \& Horticulture & 212 \\
\rowcolor{life} & Entertainment & Entertainment & 207 \\
\rowcolor{life} & & Movie \& TV Show & 230 \\
\rowcolor{life} & & Podcasting & 211 \\
\rowcolor{life} & & Hobbies & 208 \\
\rowcolor{life} & & Photography Basics & 214 \\
\midrule

\rowcolor{home} Home \& Family & Home \& Living & Home Safety & 189 \\
\rowcolor{home} & & Pet Care & 208 \\
\rowcolor{home} & & Waste Management & 207 \\
\rowcolor{home} & & Home Maintenance & 261 \\
\rowcolor{home} & Family & Parenting & 144 \\
\rowcolor{home} & & Relationships & 173 \\
\rowcolor{home} & & Teens \& Youth & 161 \\
\rowcolor{home} & Food \& Cooking & Cooking \& Recipes & 274 \\
\rowcolor{home} & & Food Safety & 219 \\
\rowcolor{home} & & Nutrition \& Diet & 151 \\
\midrule

\rowcolor{cult} Culture \& Society & Arts \& Design & Art Techniques \& Architecture & 175 \\
\rowcolor{cult} & & Interior Design & 200 \\
\rowcolor{cult} & Culture \& Religion & Cultural Awareness & 243 \\
\rowcolor{cult} & & Religious Studies & 215 \\
\rowcolor{cult} & & Holidays \& Traditions & 236 \\
\rowcolor{cult} & Legal & Legal Rights & 219 \\
\rowcolor{cult} & & Law & 206 \\
\rowcolor{cult} & Ethics \& Morality & Ethical Living & 171 \\
\rowcolor{cult} & & Ethics & 172 \\
\midrule

\rowcolor{envi} Environment & Environment & Sustainable Living & 178 \\
\rowcolor{envi} & Weather & Weather Forecasting & 205 \\
\midrule

\rowcolor{misc} Miscellaneous & Miscellaneous & Global Facts & 228 \\
\rowcolor{misc} & & News \& Information & 203 \\
\end{longtable}

\section{Mobile Relevance Score}

We calculate the Mobile-Relevance Score for all topics across different benchmarks using Equation~\ref{eq:topic_mrscore} as discussed in Section~\ref{sec:data-stat}. Our analysis reveals that Mobile-MMLU achieves the highest average relevance score of 5.88, almost twice as high as standard MMLU (3.13) and MMLU-Pro (3.00). This significant difference is further illustrated in Table~\ref{tab:stats-mrscore-overview}, where Mobile-MMLU contains 21 highly relevant topics (scoring $\geq 7.0$) and 38 moderately relevant topics (scoring 5.0-6.9), while neither MMLU nor MMLU Pro include any topics in these higher relevance categories.

The detailed breakdown in Table~\ref{tab:all-topics-by-range} demonstrates that the highest scoring topics in Mobile-MMLU are directly related to daily mobile use cases, with Travel Planning (8.43), First Aid (8.41), and Photography \& Smartphone Photography (8.05) leading the rankings. This distribution highlights how traditional benchmarks fail to adequately capture the information needs of mobile phone users, substantially reducing the practical utility of smaller language models designed for mobile applications. Our dataset successfully addresses these limitations by focusing on knowledge domains that are directly applicable to mobile users' information-seeking behavior.

\begin{table}[htbp]
\centering
\caption{Statistical Overview of Relevance Scores Across MMLU Variants.}
\label{tab:stats-mrscore-overview}
\begin{tabular}{lccccc}
\toprule
\textbf{Benchmark} & \textbf{Avg. Score} & \cellcolor{highrelevance}\textcolor{white}{\textbf{Topics $\geq$7.0}} & \cellcolor{mediumrelevance}\textcolor{white}{\textbf{Topics 5.0-6.9}} & \cellcolor{lowmediumrelevance}{\textbf{Topics 3.0-4.9}} & \cellcolor{lowrelevance}{\textbf{Topics $<$3.0}} \\
\midrule
\textbf{Mobile MMLU} & \textbf{5.88} & \cellcolor{highrelevance}\textcolor{white}{\textbf{21}} & \cellcolor{mediumrelevance}\textcolor{white}{38} & \cellcolor{lowmediumrelevance}{21} & \cellcolor{lowrelevance}{0} \\
MMLU & 3.13 & \cellcolor{highrelevance}\textcolor{white}{0} & \cellcolor{mediumrelevance}\textcolor{white}{0} & \cellcolor{lowmediumrelevance}{\textbf{29}} & \cellcolor{lowrelevance}{18} \\
MMLU Pro & 3.00 & \cellcolor{highrelevance}\textcolor{white}{0} & \cellcolor{mediumrelevance}\textcolor{white}{0} & \cellcolor{lowmediumrelevance}{6} & \cellcolor{lowrelevance}{\textbf{8}} \\
\bottomrule
\end{tabular}
\end{table}

\begin{longtable}{lcc}
\caption{Complete Relevance Scores Grouped by Relevance Range.}
\label{tab:all-topics-by-range} \\

\toprule
\textbf{Topic} & \textbf{Benchmark} & \textbf{Relevance Score} \\
\midrule
\endfirsthead

\multicolumn{3}{c}{\tablename\ \thetable{} -- Continued from previous page} \\
\toprule
\textbf{Topic} & \textbf{Benchmark} & \textbf{Relevance Score} \\
\midrule
\endhead

\midrule
\multicolumn{3}{r}{\textit{Continued on next page}} \\
\endfoot

\bottomrule
\endlastfoot

\multicolumn{3}{c}{\cellcolor{highrelevance}\textcolor{white}{\textbf{High Relevance (7.0-10.0)}}} \\
\midrule
Travel Planning & Mobile MMLU & \cellcolor{highrelevance}\textcolor{white}{8.43} \\
First Aid & Mobile MMLU & \cellcolor{highrelevance}\textcolor{white}{8.41} \\
Photography \& Smartphone Photography & Mobile MMLU & \cellcolor{highrelevance}\textcolor{white}{8.05} \\
Shopping & Mobile MMLU & \cellcolor{highrelevance}\textcolor{white}{7.96} \\
Technical Help & Mobile MMLU & \cellcolor{highrelevance}\textcolor{white}{7.88} \\
Digital Literacy & Mobile MMLU & \cellcolor{highrelevance}\textcolor{white}{7.79} \\
Mobile Customization & Mobile MMLU & \cellcolor{highrelevance}\textcolor{white}{7.74} \\
Cybersecurity & Mobile MMLU & \cellcolor{highrelevance}\textcolor{white}{7.65} \\
Online Privacy & Mobile MMLU & \cellcolor{highrelevance}\textcolor{white}{7.43} \\
Social Media & Mobile MMLU & \cellcolor{highrelevance}\textcolor{white}{7.38} \\
Automotive Care & Mobile MMLU & \cellcolor{highrelevance}\textcolor{white}{7.33} \\
Personal Finance & Mobile MMLU & \cellcolor{highrelevance}\textcolor{white}{7.32} \\
Outdoor Survival Skills & Mobile MMLU & \cellcolor{highrelevance}\textcolor{white}{7.26} \\
Mental Health & Mobile MMLU & \cellcolor{highrelevance}\textcolor{white}{7.15} \\
Cooking \& Recipes & Mobile MMLU & \cellcolor{highrelevance}\textcolor{white}{7.11} \\
Sports & Mobile MMLU & \cellcolor{highrelevance}\textcolor{white}{7.11} \\
Time Management & Mobile MMLU & \cellcolor{highrelevance}\textcolor{white}{7.05} \\
Home Safety & Mobile MMLU & \cellcolor{highrelevance}\textcolor{white}{7.06} \\
Fashion \& Style & Mobile MMLU & \cellcolor{highrelevance}\textcolor{white}{7.00} \\
Food Safety & Mobile MMLU & \cellcolor{highrelevance}\textcolor{white}{7.00} \\
Physical Fitness & Mobile MMLU & \cellcolor{highrelevance}\textcolor{white}{7.00} \\

\midrule
\multicolumn{3}{c}{\cellcolor{mediumrelevance}\textcolor{white}{\textbf{Medium Relevance (5.0-6.9)}}} \\
\midrule
Nutrition \& Diet & Mobile MMLU & \cellcolor{mediumrelevance}\textcolor{white}{6.93} \\
Pet Care & Mobile MMLU & \cellcolor{mediumrelevance}\textcolor{white}{6.85} \\
Creativity & Mobile MMLU & \cellcolor{mediumrelevance}\textcolor{white}{6.81} \\
Basic Life Skills & Mobile MMLU & \cellcolor{mediumrelevance}\textcolor{white}{6.80} \\
Social Etiquette & Mobile MMLU & \cellcolor{mediumrelevance}\textcolor{white}{6.80} \\
Parenting & Mobile MMLU & \cellcolor{mediumrelevance}\textcolor{white}{6.79} \\
Medical \& Health Knowledge & Mobile MMLU & \cellcolor{mediumrelevance}\textcolor{white}{6.78} \\
Home Maintenance & Mobile MMLU & \cellcolor{mediumrelevance}\textcolor{white}{6.73} \\
Entertainment & Mobile MMLU & \cellcolor{mediumrelevance}\textcolor{white}{6.70} \\
Weather Forecasting & Mobile MMLU & \cellcolor{mediumrelevance}\textcolor{white}{6.50} \\
Digital Detox & Mobile MMLU & \cellcolor{mediumrelevance}\textcolor{white}{6.37} \\
Relationships & Mobile MMLU & \cellcolor{mediumrelevance}\textcolor{white}{6.24} \\
E-commerce & Mobile MMLU & \cellcolor{mediumrelevance}\textcolor{white}{6.22} \\
News \& Information & Mobile MMLU & \cellcolor{mediumrelevance}\textcolor{white}{6.15} \\
Legal Rights & Mobile MMLU & \cellcolor{mediumrelevance}\textcolor{white}{6.14} \\
Teens \& Youth & Mobile MMLU & \cellcolor{mediumrelevance}\textcolor{white}{6.13} \\
Environmental \& Sustainable Living & Mobile MMLU & \cellcolor{mediumrelevance}\textcolor{white}{6.12} \\
Public Speaking & Mobile MMLU & \cellcolor{mediumrelevance}\textcolor{white}{6.00} \\
Waste Management & Mobile MMLU & \cellcolor{mediumrelevance}\textcolor{white}{6.00} \\
Gardening \& Horticulture & Mobile MMLU & \cellcolor{mediumrelevance}\textcolor{white}{5.95} \\
Interior Design & Mobile MMLU & \cellcolor{mediumrelevance}\textcolor{white}{5.80} \\
Geography & Mobile MMLU & \cellcolor{mediumrelevance}\textcolor{white}{5.76} \\
Emotional Intelligence & Mobile MMLU & \cellcolor{mediumrelevance}\textcolor{white}{5.69} \\
Marketing \& Sales Strategies & Mobile MMLU & \cellcolor{mediumrelevance}\textcolor{white}{5.69} \\
Event Planning & Mobile MMLU & \cellcolor{mediumrelevance}\textcolor{white}{5.65} \\
Ergonomics & Mobile MMLU & \cellcolor{mediumrelevance}\textcolor{white}{5.60} \\
Hobbies & Mobile MMLU & \cellcolor{mediumrelevance}\textcolor{white}{5.60} \\
Career Development & Mobile MMLU & \cellcolor{mediumrelevance}\textcolor{white}{5.38} \\
Global Facts & Mobile MMLU & \cellcolor{mediumrelevance}\textcolor{white}{5.32} \\
Holidays \& Traditions & Mobile MMLU & \cellcolor{mediumrelevance}\textcolor{white}{5.30} \\
Cultural Awareness & Mobile MMLU & \cellcolor{mediumrelevance}\textcolor{white}{5.29} \\
Personal Branding & Mobile MMLU & \cellcolor{mediumrelevance}\textcolor{white}{5.22} \\
Movie \& TV Shows & Mobile MMLU & \cellcolor{mediumrelevance}\textcolor{white}{5.22} \\
Podcasting & Mobile MMLU & \cellcolor{mediumrelevance}\textcolor{white}{5.19} \\
Communication \& Public Speaking & Mobile MMLU & \cellcolor{mediumrelevance}\textcolor{white}{5.15} \\
Business Management & Mobile MMLU & \cellcolor{mediumrelevance}\textcolor{white}{5.13} \\
Project Management & Mobile MMLU & \cellcolor{mediumrelevance}\textcolor{white}{5.12} \\
Education Techniques & Mobile MMLU & \cellcolor{mediumrelevance}\textcolor{white}{5.07} \\
\multicolumn{3}{c}{\textit{(Note: All medium relevance topics are from Mobile MMLU)}} \\

\midrule
\multicolumn{3}{c}{\cellcolor{lowmediumrelevance}{\textbf{Low-Medium Relevance (3.0-4.9)}}} \\
\midrule
Law & Mobile MMLU & \cellcolor{lowmediumrelevance}{4.95} \\
Conflict Resolution & Mobile MMLU & \cellcolor{lowmediumrelevance}{4.87} \\
Elementary Mathematics & Mobile MMLU & \cellcolor{lowmediumrelevance}{4.80} \\
Writing Skills & Mobile MMLU & \cellcolor{lowmediumrelevance}{4.70} \\
Religious Studies & Mobile MMLU & \cellcolor{lowmediumrelevance}{4.38} \\
Conceptual Physics & Mobile MMLU & \cellcolor{lowmediumrelevance}{4.37} \\
Science Fundamentals & Mobile MMLU & \cellcolor{lowmediumrelevance}{4.32} \\
Ethical Living & Mobile MMLU & \cellcolor{lowmediumrelevance}{4.29} \\
Accounting & Mobile MMLU & \cellcolor{lowmediumrelevance}{4.20} \\
High School Mathematics & Mobile MMLU & \cellcolor{lowmediumrelevance}{4.20} \\
Basic Statistics & Mobile MMLU & \cellcolor{lowmediumrelevance}{4.19} \\
Social Sciences & Mobile MMLU & \cellcolor{lowmediumrelevance}{4.10} \\
Ethics & Mobile MMLU & \cellcolor{lowmediumrelevance}{4.06} \\
Human Resources & Mobile MMLU & \cellcolor{lowmediumrelevance}{4.00} \\
Art Techniques \& Architecture & Mobile MMLU & \cellcolor{lowmediumrelevance}{3.88} \\
Linguistics & Mobile MMLU & \cellcolor{lowmediumrelevance}{3.82} \\
Political Systems & Mobile MMLU & \cellcolor{lowmediumrelevance}{3.74} \\
Reading \& Literature & Mobile MMLU & \cellcolor{lowmediumrelevance}{3.71} \\
Logical Fallacies & Mobile MMLU & \cellcolor{lowmediumrelevance}{3.66} \\
Formal Logic & Mobile MMLU & \cellcolor{lowmediumrelevance}{3.52} \\
World History & Mobile MMLU & \cellcolor{lowmediumrelevance}{3.50} \\
\midrule
Miscellaneous & MMLU & \cellcolor{lowmediumrelevance}{4.34} \\
Elementary Mathematics & MMLU & \cellcolor{lowmediumrelevance}{4.20} \\
Human Sexuality & MMLU & \cellcolor{lowmediumrelevance}{4.14} \\
Computer Security & MMLU & \cellcolor{lowmediumrelevance}{4.00} \\
Virology & MMLU & \cellcolor{lowmediumrelevance}{3.94} \\
Clinical Knowledge & MMLU & \cellcolor{lowmediumrelevance}{3.93} \\
Nutrition & MMLU & \cellcolor{lowmediumrelevance}{3.73} \\
World Religions & MMLU & \cellcolor{lowmediumrelevance}{3.68} \\
Human Aging & MMLU & \cellcolor{lowmediumrelevance}{3.67} \\
Management & MMLU & \cellcolor{lowmediumrelevance}{3.64} \\
Marketing & MMLU & \cellcolor{lowmediumrelevance}{3.60} \\
Public Relations & MMLU & \cellcolor{lowmediumrelevance}{3.42} \\
High School Computer Science & MMLU & \cellcolor{lowmediumrelevance}{3.40} \\
High School Geography & MMLU & \cellcolor{lowmediumrelevance}{3.36} \\
High School Government \& Politics & MMLU & \cellcolor{lowmediumrelevance}{3.33} \\
Security Studies & MMLU & \cellcolor{lowmediumrelevance}{3.33} \\
Global Facts & MMLU & \cellcolor{lowmediumrelevance}{3.27} \\
US Foreign Policy & MMLU & \cellcolor{lowmediumrelevance}{3.27} \\
Medical Genetics & MMLU & \cellcolor{lowmediumrelevance}{3.27} \\
High School Macroeconomics & MMLU & \cellcolor{lowmediumrelevance}{3.26} \\
Electrical Engineering & MMLU & \cellcolor{lowmediumrelevance}{3.25} \\
Anatomy & MMLU & \cellcolor{lowmediumrelevance}{3.14} \\
High School Biology & MMLU & \cellcolor{lowmediumrelevance}{3.12} \\
High School Microeconomics & MMLU & \cellcolor{lowmediumrelevance}{3.12} \\
College Medicine & MMLU & \cellcolor{lowmediumrelevance}{3.11} \\
High School Psychology & MMLU & \cellcolor{lowmediumrelevance}{3.10} \\
Professional Psychology & MMLU & \cellcolor{lowmediumrelevance}{3.09} \\
Sociology & MMLU & \cellcolor{lowmediumrelevance}{3.09} \\
Conceptual Physics & MMLU & \cellcolor{lowmediumrelevance}{3.08} \\
International Law & MMLU & \cellcolor{lowmediumrelevance}{3.08} \\
Astronomy & MMLU & \cellcolor{lowmediumrelevance}{3.06} \\
High School Statistics & MMLU & \cellcolor{lowmediumrelevance}{3.04} \\
Professional Accounting & MMLU & \cellcolor{lowmediumrelevance}{3.03} \\
Professional Medicine & MMLU & \cellcolor{lowmediumrelevance}{3.00} \\
Business Ethics & MMLU & \cellcolor{lowmediumrelevance}{3.00} \\
\midrule
Business & MMLU Pro & \cellcolor{lowmediumrelevance}{3.73} \\
Other & MMLU Pro & \cellcolor{lowmediumrelevance}{3.55} \\
Psychology & MMLU Pro & \cellcolor{lowmediumrelevance}{3.31} \\
Health & MMLU Pro & \cellcolor{lowmediumrelevance}{3.30} \\
Biology & MMLU Pro & \cellcolor{lowmediumrelevance}{3.19} \\
Economics & MMLU Pro & \cellcolor{lowmediumrelevance}{3.19} \\

\midrule
\multicolumn{3}{c}{\cellcolor{lowrelevance}{\textbf{Low Relevance ($<$3.0)}}} \\
\midrule
High School Chemistry & MMLU & \cellcolor{lowrelevance}{2.95} \\
High School Mathematics & MMLU & \cellcolor{lowrelevance}{2.93} \\
Machine Learning & MMLU & \cellcolor{lowrelevance}{2.92} \\
College Physics & MMLU & \cellcolor{lowrelevance}{2.91} \\
Moral Scenarios & MMLU & \cellcolor{lowrelevance}{2.89} \\
Prehistory & MMLU & \cellcolor{lowrelevance}{2.86} \\
Econometrics & MMLU & \cellcolor{lowrelevance}{2.83} \\
High School Physics & MMLU & \cellcolor{lowrelevance}{2.81} \\
Formal Logic & MMLU & \cellcolor{lowrelevance}{2.79} \\
Logical Fallacies & MMLU & \cellcolor{lowrelevance}{2.78} \\
Philosophy & MMLU & \cellcolor{lowrelevance}{2.76} \\
College Biology & MMLU & \cellcolor{lowrelevance}{2.75} \\
Jurisprudence & MMLU & \cellcolor{lowrelevance}{2.73} \\
College Computer Science & MMLU & \cellcolor{lowrelevance}{2.73} \\
Moral Disputes & MMLU & \cellcolor{lowrelevance}{2.68} \\
College Chemistry & MMLU & \cellcolor{lowrelevance}{2.60} \\
College Mathematics & MMLU & \cellcolor{lowrelevance}{2.55} \\
Abstract Algebra & MMLU & \cellcolor{lowrelevance}{2.55} \\
Professional Law & MMLU & \cellcolor{lowrelevance}{2.49} \\
High School US History & MMLU & \cellcolor{lowrelevance}{2.45} \\
High School European History & MMLU & \cellcolor{lowrelevance}{2.11} \\
High School World History & MMLU & \cellcolor{lowrelevance}{2.08} \\
\midrule
Computer Science & MMLU Pro & \cellcolor{lowrelevance}{2.95} \\
Mathematics & MMLU Pro & \cellcolor{lowrelevance}{2.91} \\
Physics & MMLU Pro & \cellcolor{lowrelevance}{2.84} \\
Philosophy & MMLU Pro & \cellcolor{lowrelevance}{2.80} \\
Chemistry & MMLU Pro & \cellcolor{lowrelevance}{2.63} \\
Engineering & MMLU Pro & \cellcolor{lowrelevance}{2.59} \\
History & MMLU Pro & \cellcolor{lowrelevance}{2.55} \\
Law & MMLU Pro & \cellcolor{lowrelevance}{2.45} \\
\end{longtable}

\section{Prompt for Generating Mobile Relevance Score}

\begin{tcolorbox}
You are an expert in evaluating the relevance of different types of questions for mobile devices. Assess a given question using the criteria of practical value, mobile-friendliness, and usage patterns to return a relevance score from 1 to 10.

Return a relevance score for a given question from 1 to 10 based on these criteria for mobile devices:
- Practical Value: Is the information needed in daily life or in real-world tasks? Does it solve problems or support decisions while away from a desk?

- Mobile-Friendliness: Can the content be effectively consumed on mobile (format, interactivity)?

- Usage patterns: Is this the kind of information someone would look up on their phone? Is it useful for in-the-moment decision-making?

\#\#Question\#\#

\{question\}

Return only the numeric score without explanation.
\end{tcolorbox}

\section{Topic and Category Wise Performance Variations Models}

\begin{figure}[t]
  \centering
  \includegraphics[width=0.95\linewidth]{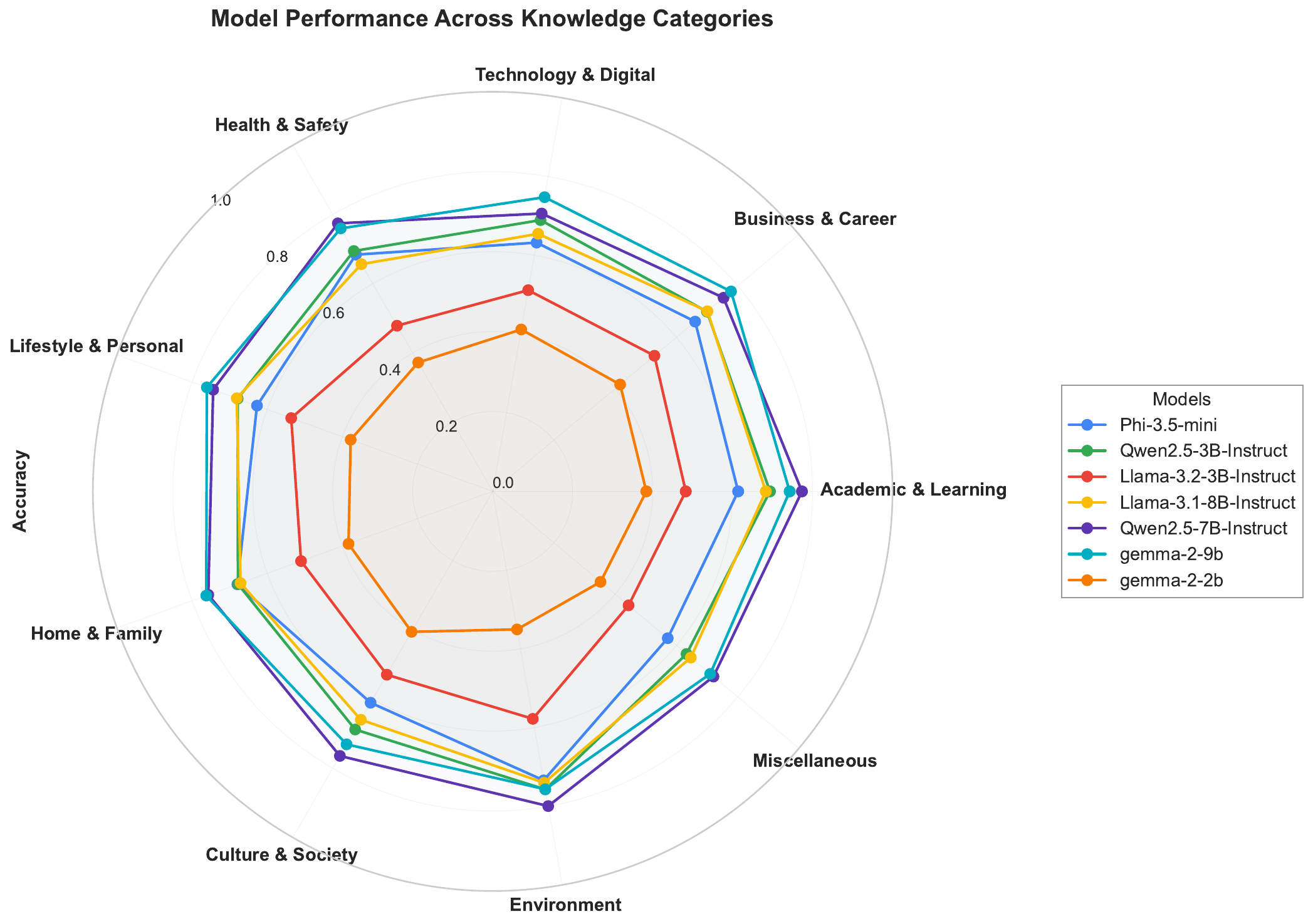}
  \caption{Radar chart showing the comparative accuracy of seven language models across nine knowledge categories relevant to mobile users.}
  \label{fig:radar_chart}
\end{figure}

\begin{figure}[t]
  \centering
  \includegraphics[width=0.95\linewidth]{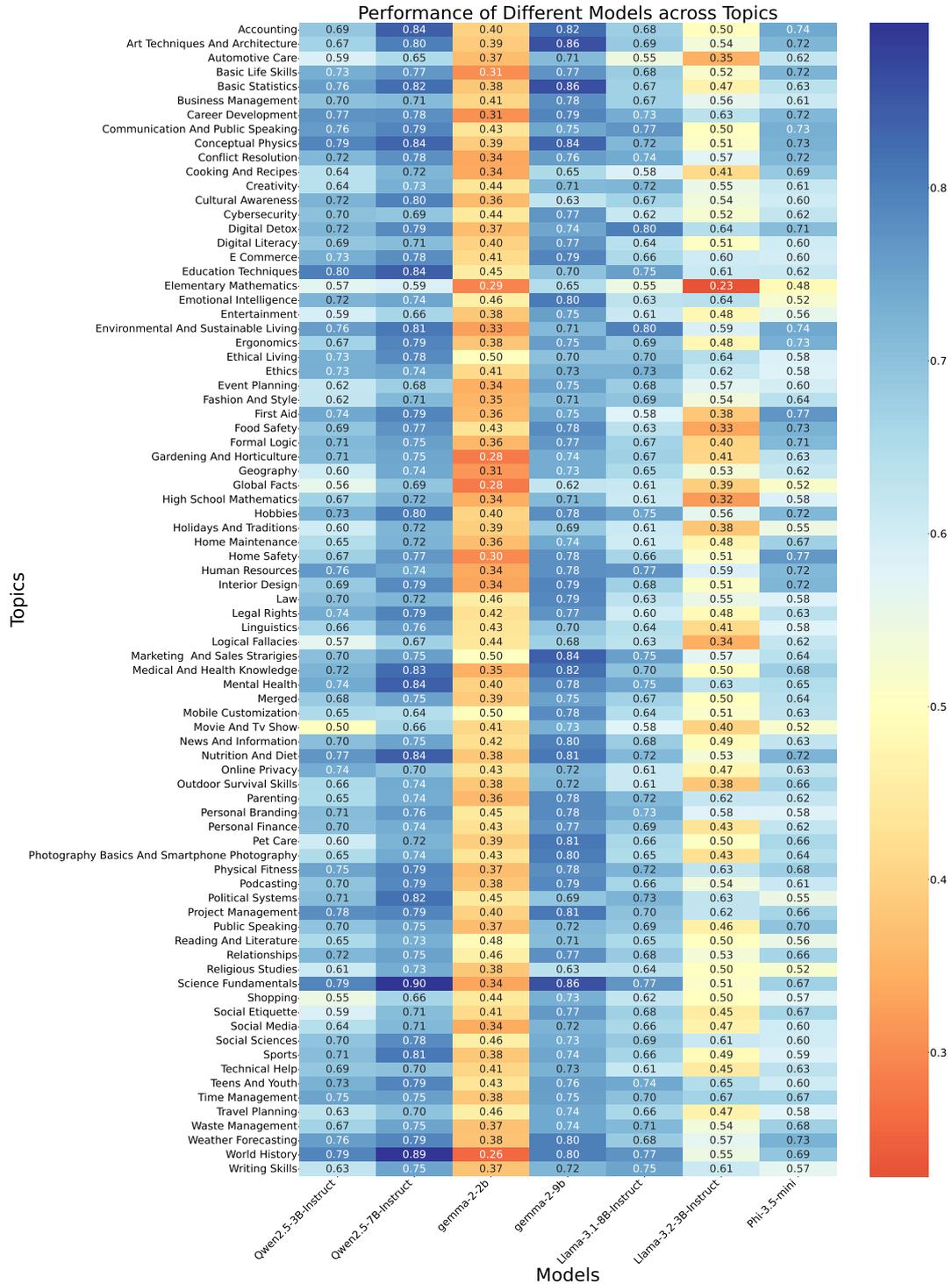}
  \caption{Heatmap visualizing the performance comparison of seven language models (Qwen2.5-3B-Instruct, Qwen2.5-7B-Instruct, Gemma-2-2b, Gemma-2-9b, Llama-3.1-8B-Instruct, Llama-3.2-3B-Instruct, and Phi-3.5-mini) across 80 topics in the Mobile-MMLU dataset. Color intensity indicates performance score, with darker blue representing higher accuracy.}
  \label{fig:heatmap_topicwise}
\end{figure}

Figure~\ref{fig:heatmap_topicwise} illustrates the comparative performance of seven language models across 80 topics in our Mobile-MMLU dataset. The heatmap reveals significant performance variations both in models and topics, highlighting the critical need for mobile-specific evaluation benchmarks. Traditional benchmarks such as MMLU and MMLU-Pro fail to adequately capture the diverse knowledge domains essential for effective mobile applications, as demonstrated by their substantially lower Mobile-Relevance Scores shown in Section~\ref{sec:data-stat}. Our topic-specific analysis provides valuable insight for selecting appropriate models for different mobile application scenarios based on their relative strengths in domains most relevant to specific use cases.

Figure~\ref{fig:radar_chart} presents a radar chart visualizing model performance in nine distinct knowledge categories critical for mobile applications. We observe that larger models generally outperform their smaller counterparts within the same family (Qwen2.5-7B vs. 3B, Gemma-2-9b vs. 2b), with Gemma-2-9b demonstrating particularly strong performance in Technology \& Digital. In contrast, the Academic \& Learning category exhibits greater variability between models, suggesting that traditional academic evaluations may not directly translate to mobile-relevant performance. These patterns confirm that mobile-specific benchmarking is essential for developing models that serve mobile users effectively.

\begin{figure}[t]
  \centering
  \includegraphics[width=0.98\linewidth]{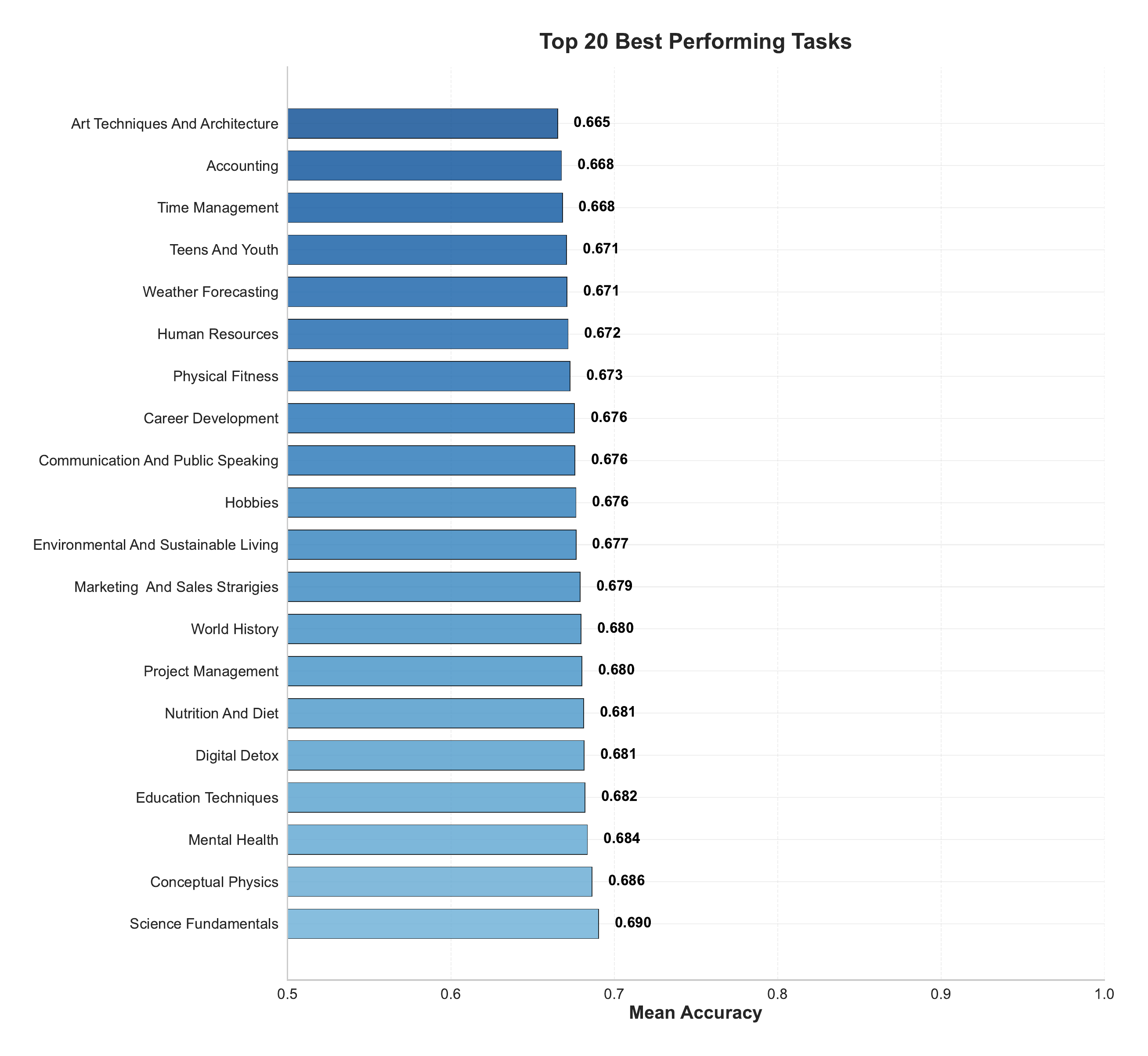}
  \caption{Top 20 best-performing topics across models in the Mobile-MMLU dataset, highlighting domains where current models show promise for mobile applications.}
  \label{fig:task_best}
\end{figure}

\begin{figure}[t]
  \centering
  \includegraphics[width=0.95\linewidth]{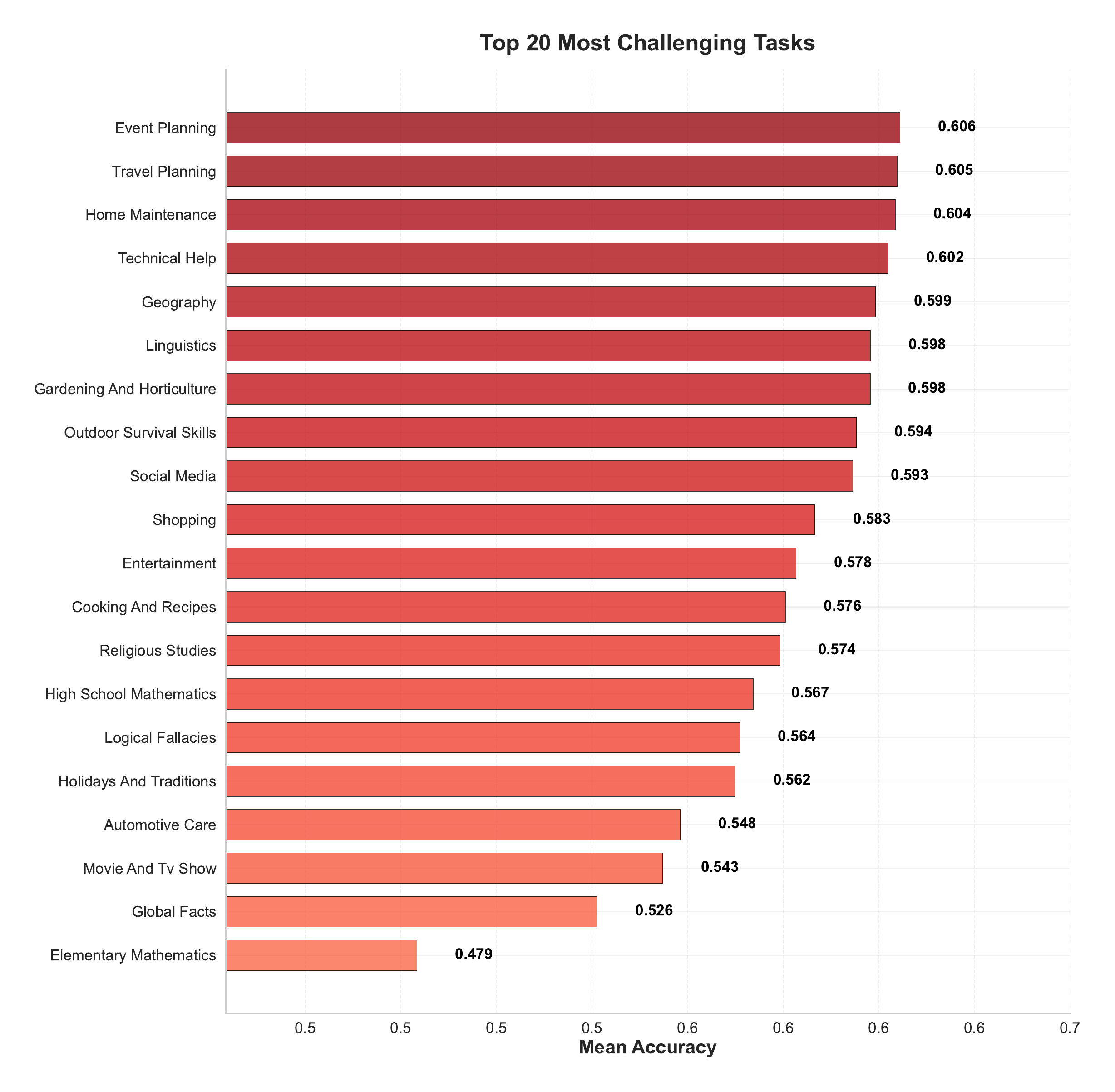}
  \caption{Top 20 most challenging topics in the Mobile-MMLU dataset, identifying critical areas for improvement in mobile-oriented language models.}
  \label{fig:task_worst}
\end{figure}

Figures~\ref{fig:task_best} and~\ref{fig:task_worst} provide additional information by identifying the easiest and most challenging topics in the Mobile-MMLU data set. Science Fundamentals achieves the highest mean accuracy (0.690) in all models, followed closely by Conceptual Physics (0.686) and Mental Health (0.684). In particular, many high-relevance mobile topics such as Digital Detox (0.681), Nutrition \& Diet (0.681), and Project Management (0.680) appear among the best performing tasks, indicating promising capabilities for practical mobile applications.
In contrast, Figure~\ref{fig:task_worst} highlights the most challenging topics, with Elementary Mathematics showing surprisingly low performance (0.479 mean accuracy). More concerning for mobile applications are the difficulties models face with highly mobile-relevant domains: Travel Planning (0.605), Technical Help (0.602), and Shopping (0.583). These are precisely the practical, everyday tasks that mobile users frequently need assistance with, yet they pose significant challenges for current models. Such findings would remain hidden when using conventional benchmarks like MMLU and MMLU-Pro, which lack sufficient coverage of these mobile-critical domains. These results validate the contribution of Mobile-MMLU as an essential evaluation framework to develop more capable and useful language models for mobile devices, addressing the practical needs that traditional benchmarks overlook.

\section{Answer Order Sensitivity}

\begin{figure}[ht]
  \centering
  \includegraphics[width=0.95\linewidth]{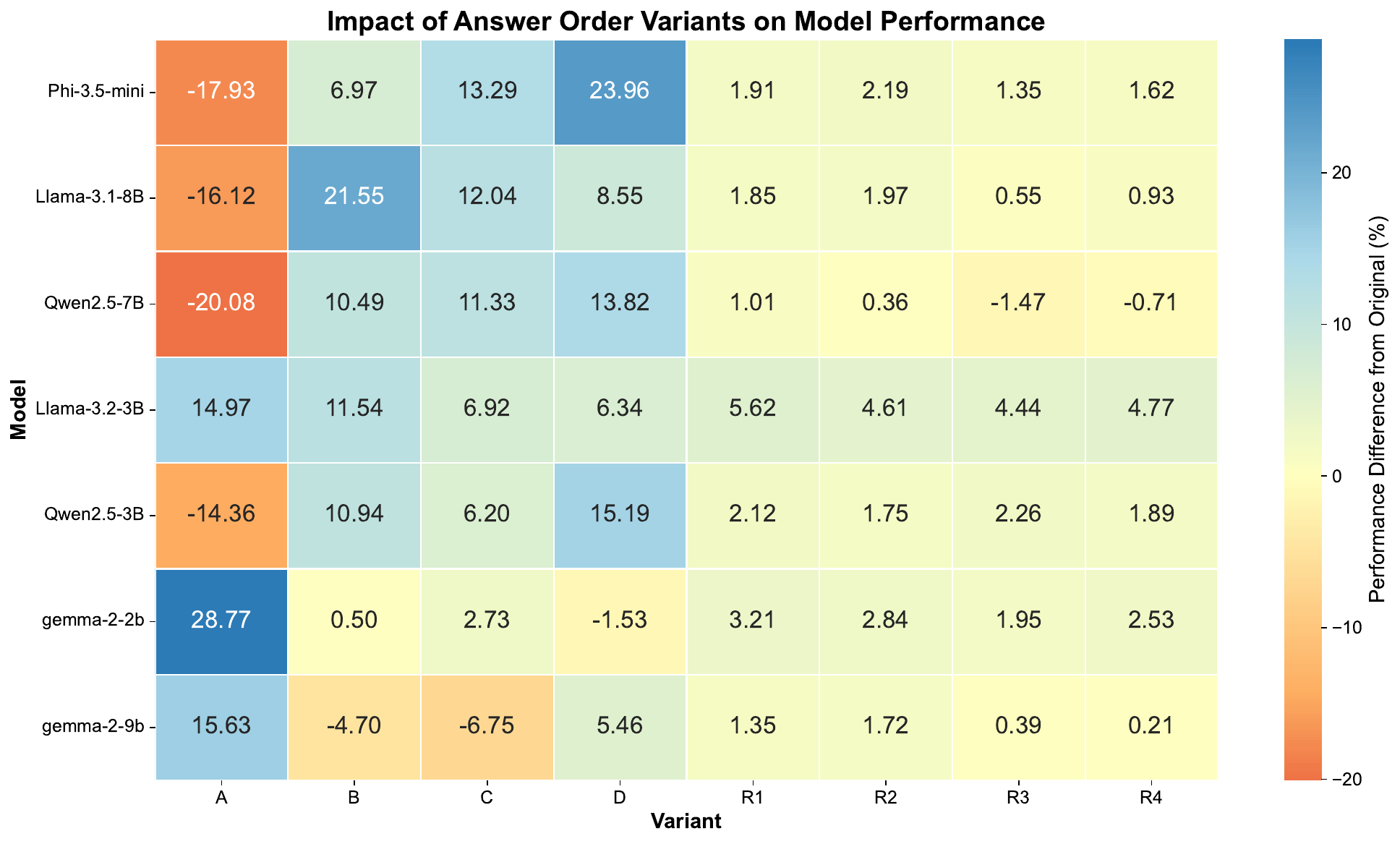}
  \caption{Impact of answer order variant in the model performance. Heatmap shows the performance difference from original model in terms of percentage. Note that, Variations because of different options shows the option led position bias in different model. However we also observe that in case of random options, all variants have almost similar performance which shows the robustness of our benchmark.}
  \label{fig:variant_heatmap}
\end{figure}

\begin{figure}[ht]
  \centering
  \includegraphics[width=0.95\linewidth]{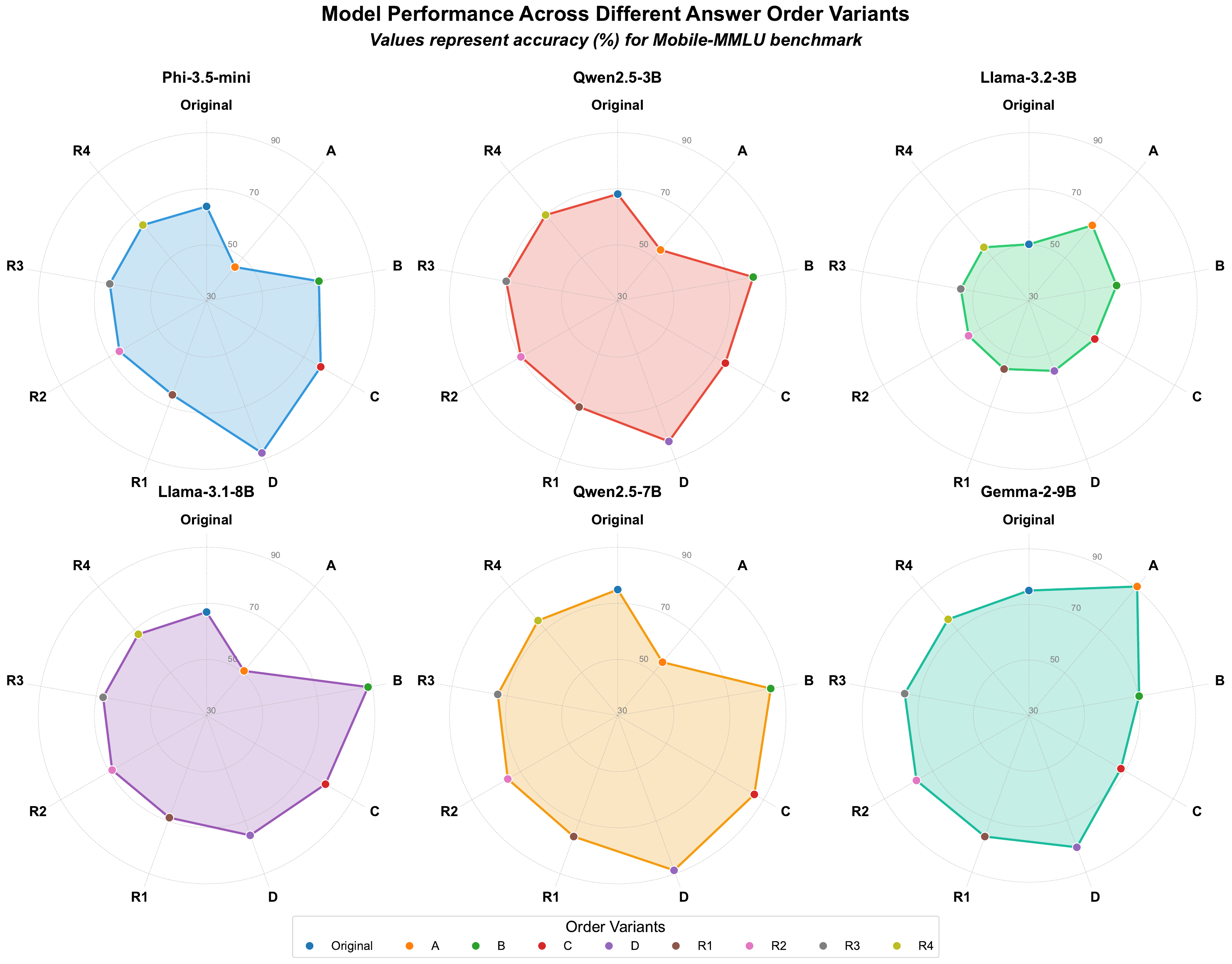}
  \caption{Answer Order Sensitivity Across Different Language Models: Impact of answer order variant in the model performance. Radar plot shows the performance for different options, different random orders and different models.}
  \label{fig:model_radar_subplots}
\end{figure}

Figures~\ref{fig:variant_heatmap} and~\ref{fig:model_radar_subplots} illustrate how model performance varies with different answer orderings. Our analysis reveals two significant patterns in this sensitivity testing. First, when examining the randomized variants (R1-R4), we observe minimal performance fluctuations across all models (typically within $\pm 3\%$), indicating that the benchmark maintains consistency when answer positions are randomized. This stability suggests that Mobile-MMLU provides reliable and robust evaluation when using randomized answer positions.

Second, we find substantial performance variations when correct answers are fixed in specific positions (A, B, C, D). For instance, Gemma-2-2b shows a 28.77\% improvement when correct answers appear in position A, while Phi-3.5-mini demonstrates a 23.96\% improvement with position D. Conversely, several models show performance degradation with certain positions—notably Qwen2.5-7B (-20.08\%) and Phi-3.5-mini (-17.93\%) with position A. These pronounced variations reveal positional biases inherent in different model architectures. 

The consistent performance across randomized variants compared to high variation with fixed positions demonstrates that position bias significantly impacts model evaluation. This finding has important implications for benchmark design, suggesting that fixed-position evaluations may artificially inflate or depress model scores based on their inherent positional preferences rather than their actual knowledge capabilities. Stable performance with randomized answer orders confirms that Mobile-MMLU provides a more objective and robust evaluation.

\end{document}